\documentclass[12pt]{article}
\RequirePackage[OT1]{fontenc} \RequirePackage{amsthm,amsmath}
\usepackage{graphics,epsfig,epsf,psfrag}
\usepackage{url}
\usepackage{amsmath,amsfonts,amssymb,amsthm}
\usepackage{bbm, bm}
\usepackage{graphicx,lscape,rotate}
\usepackage{dashrule} % To make dotted/dashed lines
\usepackage{epstopdf}
\usepackage{algorithmic, algorithm}
\usepackage{enumerate, framed}
\usepackage{colortbl}
\usepackage{color, appendix}
\usepackage[round]{natbib}
\RequirePackage[colorlinks,citecolor=blue,urlcolor=blue]{hyperref}
\usepackage{pstricks,pst-node,pst-tree}  % PLotting Trees
\usepackage{graphics, hyperref}  % <-- special graphics package

\let\BLS=\baselinestretch
\makeatletter
\newcommand{\singlespacing}{\let\CS=\@currsize\renewcommand{\baselinestretch}{1}\small\CS}
\newcommand{\doublespacing}{\let\CS=\@currsize\renewcommand{
\baselinestretch}{1.5}\small\CS}
\newcommand{\normalspacing}{\let\CS=\@currsize\renewcommand{\baselinestretch}{\BLS}\small\CS}
\makeatother

\providecommand{\keywords}[1]{\textbf{\textit{Keywords:~}} #1}

\newtheorem{proposition}{Proposition}[section]
\newtheorem*{remark*}{Remark} %For having unnumbered remarks.
\theoremstyle{definition}
\newtheorem{definition}{Definition}%[section]

\newcommand{\x}{\mathbf{x}}

\newcommand{\expit}{\mbox{expit}}

\DeclareMathOperator*{\argmax}{\arg\!\max}
\newcommand\ind{\protect\mathpalette{\protect\indT}{\perp}}
\def\indT#1#2{\mathrel{\rlap{$#1#2$}\mkern2mu{#1#2}}}

\newcount\colveccount
\newcommand*\colvec[1]{
        \global\colveccount#1
        \begin{pmatrix}
        \colvecnext
}
\def\colvecnext#1{
        #1
        \global\advance\colveccount-1
        \ifnum\colveccount>0
                \\
                \expandafter\colvecnext
        \else
                \end{pmatrix}
        \fi
}

\makeatletter
\newcommand{\Spvek}[2][r]{%
  \gdef\@VORNE{1}
  \left(\hskip-\arraycolsep%
    \begin{array}{#1}\vekSp@lten{#2}\end{array}%
  \hskip-\arraycolsep\right)}

\def\vekSp@lten#1{\xvekSp@lten#1;vekL@stLine;}
\def\vekL@stLine{vekL@stLine}
\def\xvekSp@lten#1;{\def\temp{#1}%
  \ifx\temp\vekL@stLine
  \else
    \ifnum\@VORNE=1\gdef\@VORNE{0}
    \else\@arraycr\fi%
    #1%
    \expandafter\xvekSp@lten
  \fi}
\makeatother

\begin{document}

% \doublespacing % <-- if you want to doublespace your document

\title{Random Forests of Interaction Trees for Estimating Individualized Treatment Effects in Randomized Trials}
\author{\textbf{Xiaogang Su}, \textbf{Annette T.~Pe\~{n}a} \\
Department of Mathematical Sciences \\
University of Texas, El Paso, TX 79968 \vspace{.1in} \\
\textbf{Lei Liu} \\
Department of Preventive Medicine \\
Northwestern University, Chicago, IL 60611 \vspace{.1in} \\
and \textbf{Richard A.~Levine} \\
Department of Mathematics and Statistics \\
San Diego State University, San Diego, CA 92182
}

% \date{\today}
\date{}
\maketitle

\renewcommand{\abstractname}{\large Abstract \vspace{.1in}}
\begin{abstract}
{\normalsize Assessing heterogeneous treatment effects has become
a growing interest in advancing precision medicine. Individualized treatment effects (ITE) play a critical role in
such an endeavor. Concerning experimental data collected from
randomized trials, we put forward a method, termed random forests
of interaction trees (RFIT), for estimating ITE on the basis of
interaction trees \citep{Su:2009}. To this end, we first propose a
smooth sigmoid surrogate (SSS) method, as an alternative to greedy
search, to speed up tree construction. RFIT outperforms the
traditional `separate regression' approach in estimating ITE.
Furthermore, standard errors for the estimated ITE via RFIT can be
obtained with the infinitesimal jackknife method. We assess and
illustrate the use of RFIT via both simulation and the analysis of
data from an acupuncture headache trial.}
\end{abstract}

\keywords{Individualized treatment effects, Infinitesimal jackknife, Precision medicine, Random forests, Treatment-by-covariates interaction.}

\section{Introduction}
Precision medicine aims to optimize the
delivery of stratified or individualized therapies by integrating comprehensive patient data. This emerging approach has become a growing interest
in many biomedical applications. To advance precision medicine, it
is crucial to understand the differential effects of a treatment
as opposed to its overall main effect in the conventional practice
of  medical decisions.

There are many approaches in this endeavor; see
~\citet{Lipkovich:2017} for a recent survey. Among them,
tree-based methods are dominant for several reasons. Built simply
on the basis of a two-sample test statistic, trees facilitate a
powerful comprehensive modeling by recursively grouping data.
Differential treatment effects essentially involve
treatment-by-covariates interactions, which may be of nonlinear
forms and of high orders. Trees excel in dealing with complex
interactions. Tree models are capable of handling high-dimensional
covariates of mixed types and an off-the-shelf tool in the sense
that minimal data preparation is required.

Interaction trees (IT; \citeauthor{Su:2009}, \citeyear{Su:2009})
extend tree procedures to subgroup analysis by explicitly
assessing the treatment-by-covariate interaction.
\citet{Foster:2011} identifies subgroups by estimating the
potential outcomes, which they rebranded as `virtual twins'.
Another approach, SIDES (Subgroup Identification based on
Differential Effect Search) proposed by \citet{Lipkovich:2011},
addresses issues such as subgroups with enhanced treatment
effects, taking into account both efficacy and toxicity. QUINT
(QUalitative INteraction Trees; \citeauthor{Dusseldorp:2014},
\citeyear{Dusseldorp:2014}) focuses on qualitative interactions.
\citet{Loh:2015} proposes a tree procedure for identifying
subgroups that is less prone to biased variable selection. The
optimal treatment regime \citep{Murphy:2003} offers an alternative
way of looking at the problem. Along this direction, tree-based
approaches are also common; see, e.g., \citep{Zhang:2012} and
\citep{Laber:2015}.

There are typically two types of precision medicine: stratified
medicine and personalized medicine. The aforementioned methods
belong to the former scope, with focus on stratified treatment
effects or regimes where groups of individuals showing homogeneous
treatment effects are sought. That said, individualized treatment
effects (ITE) are of key importance in deploying tailored
treatment plans as part of personalized medicine. The ITE also
affords deeper study of treatment efficacy. Furthermore, ITE
estimation is a necessary first step for a number of methods used
in stratified medicine and optimal treatment regime; see, e.g.,
\citet{Foster:2011}, \citet{Zhang:2012}, and \citet{Laber:2015}.
% In particular, optimal treatment regimen would be trivial
% once ITE can be accurately estimated.

The focus of this article is on the estimation of ITE with data
collected from randomized trials. We examine an ensemble learning
approach that we coin as RFIT for random forests on the basis of
interaction trees \citep{Su:2009}. Our methodological contribution
is twofold: first, we introduce a faster alternative splitting
method, called smooth sigmoid surrogate (SSS), to speed up IT;
second, we extend the infinitesimal jackknife method
\citep{Efron:2014} to compute the standard errors for ITE
estimates. Moreover, we compare our proposed approach to the
commonly applied separate regression (SR). RFIT is superior to SR by working on an easier problem. We demonstrate the outperformance of RFIT over SR in estimating ITE via extensive numerical experiments.

The remainder of the article is organized as follows. In Section
\ref{sec-RFIT}, we first introduce the concept of ITE within
Rubin's causal model framework. RFIT with SSS splitting for
estimating ITE and the standard error formula for estimated ITE
are then presented in detail. Section \ref{sec-simulation}
contains simulation experiments that are designed to investigate
the performance of SSS in splitting data, to compare RFIT with the
conventional separate regression approach, and to demonstrate the
validity of the SE formula. We illustrate our proposed RFIT
approach with data from an acupuncture headache trial in Section
\ref{sec-example}.

\section{Random Forests of Interaction Trees (RFIT)}
\label{sec-RFIT}

Concerning randomized trials, consider data $\mathcal{D} = \{(y_i,
T_i, \mathbf{x}_i): i = 1, \ldots, n\}$ consisting of $n$ IID
copies of $(Y, T, \mathbf{X}),$ where $y_i$ is the continuous
response or outcome for the $i$-th subject, $T_i$ is the binary
treatment assignment indicator: 1 for the treated group and 0 for
control, and $\mathbf{x}_i = (x_{i1}, \ldots, x_{ip})^T  \in
\mathbb{R}^p$ is a $p$-dimensional covariate vector of mixed
types.

The Neyman--Rubin causal model (see, e.g.,
\citeauthor{Neyman:1923}, \citeyear{Neyman:1923};
\citeauthor{Rubin:1974}, \citeyear{Rubin:1974} \&
\citeyear{Rubin:2005}) provides a way of finely calibrating the
causal effect of treatment $T$ on the response via the concept of
potential outcomes. Let $Y'_{1}$ and $Y'_{0}$ denote the response
values for a subject when assigned to the treated and the control
group, respectively. Either $Y'_{1}$ or $Y'_{0}$, but not both,
can be observed. The observed outcome is given by $Y = Y'_1 \, T +
Y'_0 \, (1-T).$ Within this framework, the treatment effect can be
evaluated at three levels: the population level $E(Y'_{1} -
Y'_{0})$, the subpopulation level $E(Y'_{1} - Y'_{0} \, | \,
\mathbf{X} \in A)$ for a subset $A \subset \mathbb{R}^p,$ and the
unit or subject level $Y'_{1} - Y'_{0}.$ These three levels form a
hierarchy of causal inference in increasing order of strength, in
the sense that a lower-level inference can be obtained from that
of an upper-level inference, but not vice versa. Let $\delta$ be a
generic notation for treatment effect.
\begin{definition}
The \textit{individualized treatment effect} (ITE) is defined as
$\delta(\mathbf{x}) = E(Y'_1 - Y'_0 \, | \,
\mathbf{X}=\mathbf{x}).$
\end{definition}
Note that $\delta(\mathbf{x})$ is different from the (random)
unit-level effect $(Y'_{1} - Y'_{0})$. Strictly speaking,
$\delta(\mathbf{x})$ is a subpopulation-level effect among
individuals with $\mathbf{X}=\mathbf{x}.$ Nevertheless,
$\delta(\mathbf{x})$ is the finest approximation to the unit-level
inference that is possibly available in practice.

Causal inference is essentially concerned with estimating $\delta$
at different levels through the available data $\mathcal{D}.$ The
difficulty in causal inference stems primarily from the convoluted
roles (e.g., confounder, effect modifier or moderator, or
mediator) played by each covariate in $\mathbf{X}.$ For
experimental data from trials with random treatment assignment
mechanisms, $T$ is independent of other variables. As a result,
the unconfoundedness condition $ (Y_1, Y_0) \ind T \, | \,
\mathbf{X}$ \citep{Rubin:2005}, being sufficient for obtaining
population-level inference from $\mathcal{D}$, is trivially met.
Randomization renders the confounding issue of little concern;
however, covariate modification to the treatment effects remains
across both the subpopulation and unit levels.

Interaction trees (IT; \citeauthor{Su:2009}, \citeyear{Su:2009})
seek subgroups with heterogeneous treatment effects by following
the paradigm of CART \citep{Breiman:1984}; hence IT supplies
causal inference at the subpopulation level. Nevertheless, results
from IT can be building blocks for inferences at other levels: one
has the flexibility to move backward to the overall effect
estimation by integration and move forward to ITE via ensemble
learning. The main objective of this article is to examine the use
of random forests of interaction trees (RFIT) in estimating
$\delta(\mathbf{x}).$ Random forests \citep{Breiman:2001} is an
ensemble learning method, constructing a collection of tree models
and integrating results across the tree models. Among its many
merits, RF is among the top-performers in predictive modeling and
provides many useful features such as proximity matrix, variable
importance ranking, and partial dependence plot \citep{Liaw:2002}.

\subsection{SSS for Identifying the Best Cutoff Point}
\label{sec-SSS}

To extend random forests on the basis of interaction trees, one
essential ingredient is the splitting statistics. IT bisects data
by maximizing the difference in treatment effects between two
child nodes. A split on data is induced by a binary variable of
general form $\Delta = \Delta(X_j; c) = I(X_j \leq c)$ that
applies a threshold on covariate $X_j$ at cutoff point $c$; recall
that nominal variables can be made ordinal by sorting its levels
according to the treatment effect estimate at each level (see Appendix
A of \citeauthor{Su:2009}, \citeyear{Su:2009}). Any binary split $s$
results in the following $2\times 2$ table, where $n_{1L}$ denotes
the number of treated subjects in the left child node;
$\bar{y}_{1L}$ denotes the sample mean response for treated
subjects in the left child node; and similarly for other
notations.
\renewcommand{\tabcolsep}{8pt}
\renewcommand{\arraystretch}{1.0}
\begin{center}
\begin{tabular}{cccc} \hline \hline
& \multicolumn{2}{c}{Child Node} & \\
\cline{2-3} Treatment & Left  & Right  &  \\ \hline
0 & ($\bar{y}_{0L}$, $n_{0L}$) & ($\bar{y}_{0R}$, $n_{0R}$) \\
1 & ($\bar{y}_{1L}$, $n_{1L}$) & ($\bar{y}_{1R}$, $n_{1R}$) \\
\hline
\end{tabular}
\end{center}
The splitting statistic in IT can be based on the Wald test
for $H_0:~ \beta_3 =0$ in the interaction model:
\begin{equation} \label{model-interaction}
y_i = \beta_0 + \beta_1 T_i + \beta_2 \Delta_{i} + \beta_3 T_i
\cdot \Delta_{i} + \varepsilon_i \mbox{~with~}  \varepsilon_i
\stackrel{IID}{\sim} N(0, \sigma^2),
\end{equation}
where $\Delta_i = \Delta(x_{ij}; c).$ The least squares estimate of $\beta_3$ is given by $\hat{\beta_3} = (\bar{y}_{1L} -
\bar{y}_{0L}) - (\bar{y}_{1R} -  \bar{y}_{0R})$, corresponding to the concept of `difference in differences' (DID). The resultant Wald test statistic amounts to
\begin{equation}
 Q(c) = \frac{ \left\{(\bar{y}_{1L} -
\bar{y}_{0L}) - (\bar{y}_{1R} -
  \bar{y}_{0R})\right\}^2}{\hat\sigma^2 ( 1/n_{1L} + 1/n_{0L} + 1/n_{1R} +
  1/n_{0R})},
\label{Q-GS}
\end{equation}
where
\begin{equation}
\label{sigma2-hat} \hat{\sigma}^2 = \frac{1}{n-4}
\left(\sum_{i=1}^n y_i^2 - \sum_{k= 0, 1} \, \sum_{t \in \{L, R\}}
n_{kt} \, \bar{y}_{kt}^2 \right)
\end{equation}
is the pooled estimator of $\sigma^2.$ $Q(c)$ measures the
difference in treatment effects between two child nodes. With the
conventional greedy search (GS) approach, the best cutoff point
$\hat{c}$ for $X_j$ is $ \hat{c} = \argmax_{c} Q(c).$ It is worth
noting that minimizing the least squares (LS) criterion with Model
(\ref{model-interaction}) does not serve well in IT. A cutoff
point can yield the minimum LS criterion merely for its strong
additive effect (i.e., associated with $\beta_2$).

GS evaluates the splitting measure at every possible cutoff point
for $X_j$. This can be slow when the number of cutoff points to be
evaluated is large, even though GS can be implemented by updating
the computation of $Q(c)$ for neighboring $c$'s. Furthermore, this
discrete optimization procedure yields erratic measures, as
exemplified by the orange line in Figure \ref{fig01}(b). As a
result, GS may mistakenly select a local spike due to large
variation. These deficiencies motivate us to consider a smooth
alternative to GS. Noting that the discreteness steps from the
threshold indicator function $\Delta_i$ involved in many
components of the splitting statistics, our approach is to
approximate $\Delta_i$ with a smooth sigmoid function. For this
reason, we call the method `smooth sigmoid surrogate' or SSS in
short. While many sigmoid functions can be used, it is natural to
consider the logistic or expit function
\begin{equation}
s(x; a, c) = [1 + \exp\{-a(x-c) \} ]^{-1} = \frac{\exp\{a\,
(x-c)\}}{1+ \exp\{a\, (x-c)\}}, \label{expit}
\end{equation}
where $c$ is the cutoff point and $a >0$ is a shape or scale
parameter. Figure \ref{fig01}(a) depicts the expit function at
$c=0$ for different $a$ values.

To approximate $Q(c)$, we start with approximating $n_{l \tau}$
with $\tilde{n}_{lt}$ for $l=0,1$ and $t=\{L, R\}$ as follows:
\begin{equation*}
\left\{ \arraycolsep=5pt\def\arraystretch{1.2}
\begin{array}{lcl}
n_{1L} = \sum_{i=1}^n T_i  \Delta_{i}  &~\approx~& \tilde{n}_{1L} = \sum_{i=1}^n T_i s_i,   \\
n_{1R} = n_1 - n_{1L} &~\approx~& \tilde{n}_{1R} = n_1 - \tilde{n}_{1L}, \\
n_{0L} = \sum_{i=1}^n \left(1-T_i\right)  \delta_{i}  &~\approx~& \tilde{n}_{0L} = \sum_{i=1}^n (1-T_i) s_i, \\
n_{0R} = n_0 - n_{0L} &~\approx~& \tilde{n}_{0R} = n_0 - \tilde{n}_{0L},
\end{array} \right.
\end{equation*}
where $s_i = s(x_{ij}; a, c)$ approximates $\Delta_i$, $n_1 =
\sum_i T_i $ is the total number of treated individuals, and $n_0=
\sum_{i=1}^n (1-T_i)$ is the total number of untreated
individuals. Let $S_{lt} = \sum_{i\in t ~\&~ T_i=l} y_i$ denote
the associated sum of responses values, which can be approximated
in a similar manner:
\begin{equation*}
\left\{ \arraycolsep=4pt\def\arraystretch{1.2}
\begin{array}{lcl}
S_{1L} = \sum_{i=1}^n y_i T_i  \Delta_{i}  &~\approx~& \tilde{S}_{1L} = \sum_{i=1}^n y_i T_i s_i,  \\
S_{1R} = S_1 - S_{1L}  &~\approx~& \tilde{S}_{1R} = S_1 - \tilde{S}_{1L}, \\
S_{0L} = \sum_{i=1}^n y_i \left(1-T_i\right)  \Delta_{i}  &~\approx~& \tilde{S}_{0L} = \sum_{i=1}^n y_i \left(1-T_i\right) s_i, \\
S_{0R} = S_0 - S_{0L}  &~\approx~& \tilde{S}_{0R} = S_0 - \tilde{S}_{0L},
\end{array} \right.
\end{equation*}
where $S_1 = \sum_i T_i y_i$ is the sum of response values for all
treated individuals and similarly $S_0$ for the untreated. Note
that quantities $n_1$, $n_0 = n-n_1$, $S_1$, and $S_0 = \sum_i y_i
- S_1$ do not involve the split variable $\Delta_i$ and can be
computed beforehand. It follows that $ \bar{y}_{lt} = S_{lt}/n_{lt} ~\approx~
\tilde{S}_{lt}/\tilde{n}_{lt} \, = \, \tilde{y}_{lt}$ for
$l=0,1$ and $t=\{L, R\}$. Next, bringing $(\tilde{n}_{lt},
\tilde{y}_{lt})$ into (\ref{sigma2-hat}) yields its approximation
$\tilde{\sigma}^2.$ Finally, plugging all the approximated
quantities into $Q(c)$ in (\ref{Q-GS}) yields
\begin{equation}
\widetilde{Q}(c) = \frac{ \left\{(\tilde{y}_{1L} - \tilde{y}_{0L})
- (\tilde{y}_{1R} -
  \tilde{y}_{0R})\right\}^2}{\tilde\sigma^2 ( 1/ \tilde{n}_{1L} + 1/
  \tilde{n}_{0L} + 1/\tilde{n}_{1R} +
  1/\tilde{n}_{0R})}.
 \label{Q-SSS}
\end{equation}
Now $\widetilde{Q}(c)$ is a smooth objective function for $c$ only
and can be directly maximized to obtain the best cutoff point
$\hat{c}.$

Besides $c$, there is a scale parameter $a$ involved in
$\widetilde{Q}(c)$ given by (\ref{Q-SSS}). As shown by simulation
in Section \ref{sec-simulation}, the performance of the SSS method
is quite robust with respect to the choice of $a$ for a wide
range of values. Thus $a$ can be fixed \textit{a priori}. In order
to do so, we standardize the predictor $x_{ij}: = (x_{ij} -
\bar{x}_j) / \hat{\sigma}_j$, where $(\bar{x}_j, \hat{\sigma}_j)$
denote the sample mean and standard deviation of variable $X_j$,
respectively. For standardized covariates, we recommend fixing $a$
a value in [10, 50]. With fixed $a$, the best cutoff point
$\hat{c}$ can be obtained by maximizing $\tilde{Q}(c)$ with
respect to $c$ and then transformed back to the original data
scale for interpretability. This is a one-dimensional smooth
optimization problem, which can be conveniently solved by many
standard optimization routines. We use the \citet{Brent:1973}
method available in the R \citep{R:2017} function
\texttt{optimize} in our implementation. Given the nonconcave nature of the maximization problem, further techniques such as multi-start or partitioning the search range may be used in combination with Brent's method. However, as shown in our numerical studies, a plain application of Brent's method, without further efforts for locating the global optimum, works quite effectively in estimating the cutpoint.

SSS smooths out local spikes in GS splitting measures and hence
helps signify the true cutoff point; see Figure \ref{fig01}(b) for
one example. Our simulation in Section \ref{sec-simulation} shows
that SSS outperforms GS in most scenarios, especially when dealing
with weak signals. Another immediate advantage of SSS over GS is
computational efficiency. The following proposition provides an
asymptotic quantification of the computational complexity involved
in both GS and SSS splitting.
\begin{proposition} \label{prop-computational-complexity}
Consider a typical data set of size $n$ in the interaction tree
setting, where both GS and SSS are used to find the best cutoff
point $\hat{c}$ for a continuous predictor $X$ with $O(n)$
distinct values. In terms of computation complexity, GS is at best
$O\{\ln(n) \, n\}$ with the updating scheme and $O(n^2)$ without
the updating scheme while SSS is $O(mn)$, where $m$ is the number
of iterations in Brent's method.
\end{proposition}
A proof of Proposition \ref{prop-computational-complexity} is
relegated to the Supplementary Materials. Implementation of tree
methods benefits from incremental updating; see, e.g.,
\citet{LeBlanc.1993} and \citet{Utgoff:1997}. However, it is a
common wrong impression that the GS splitting with updating is
only of order $O(n).$ Updating the IT splitting statistic entails
sorting the response values according to the $X$ values within
both treatment groups. It turns out that this sorting step would
dominate the algorithm in complexity asymptotically with a rate of
$O\{\ln(n) \, n\}.$ Comparatively, SSS depends on the number of
iterations in Brent's method, $m.$ Although the number of
iterations is affected by the convergence criterion and the
desired accuracy, $m$ is generally small since Brent's method has
guaranteed convergence at a superlinear rate. Based on our limited
numerical experience, $m$ rarely gets over 15 even for large $n$.
In other words, the $O(mn)$ rate for SSS essentially amounts to
the linear rate $O(n).$

\subsection{Estimating ITE via RFIT}
\label{sec-ITE-RFIT} RFIT follows the standard paradigm of random
forests \citep{Breiman:2001}. Take a bootstrap sample
$\mathcal{D}_b$ from data $\mathcal{D}$ and then construct an IT
using $\mathcal{D}_b.$ To split a node, a subset of $m$ covariates are
randomly selected and the best cut for each covariate is
identified and compared to determine the best split of data. The
step is iterated till a large tree $\mathcal{T}_b$ is grown. Each
terminal node $\tau$ in $\mathcal{T}_b$ is summarized with an
estimated treatment effect $\hat{\delta}_{\tau_b}$, which is
simply the difference in mean response between treated and
untreated individuals falling into $\tau$, i.e.,
$$ \hat{\delta}_{\tau_b} = \sum_{i:~ \mathbf{x}_i \in \mathcal{D}_b
\cap \tau_b} \left\{ \frac{T_i y_i}{n_{1\tau_b}} - \frac{(1-T_i)
y_i}{ n_{0\tau_b}} \right\},$$ where $n_{1\tau_b} = \sum_{i:
\mathbf{x}_i \in \mathcal{D}_b \cap \tau_b} T_i$ is the number of
treated individuals in $\mathcal{D}_b$ that fall into $\tau$ and
$n_{0\tau_b}$ for the untreated.

The entire tree construction procedure is then repeated on a
number of $B$ bootstrap samples, which results in a sequence of
bootstrap trees $\{\mathcal{T}_b: b=1, 2, \ldots, B\}.$ For each
tree $\mathcal{T}_b$, an individual with covariate vector
$\mathbf{x}$ would fall into one and only one of its terminal
node, which we denote as $\tau_b(\mathbf{x}).$ Letting $
\hat{\delta}_b(\mathbf{x}) = \hat{\delta}_{\tau_b(\mathbf{x})},$
the ITE for this individual can then be estimated as
\begin{equation}
\hat{\delta}(\mathbf{x}) = \frac{1}{B} \sum_{i=1}^B
\hat{\delta}_b(\mathbf{x}). \label{ITE-RFIT}
\end{equation}

\citet{Efron:2014} discussed methods for computing standard errors
for bootstrap-based estimators  and advocated the use of
infinitesimal jackknife (IJ). The IJ approach is found preferable
in random forests, as further explored by \citet{Wager:2014}.
Proposition \ref{prop-SE} applies the IJ method to obtain a
standard error formula for estimated ITE
$\hat{\delta}(\mathbf{x}).$ Its proof is outlined in the
Supplementary Materials.
\begin{proposition}
\label{prop-SE} The IJ estimate of variance of $\hat{\delta}(\x)$
is given by
\begin{equation} \label{SE-IJ}
\hat{V} = \sum_{i=1}^n  \bar{Z}_i^2,
\end{equation}
where $\bar{Z}_i = \sum_{b=1}^B Z_{bi}/B$ and $Z_{bi} =(N_{bi}
-1)\{ \hat{\delta}_b(\x) - \hat{\delta}(\x)\}$ with $N_{bi}$ being
the number of times that the $i$-th observation appears in the
$b$-th bootstrap resample. In other words, the quantity
$\bar{Z}_i$ is the bootstrap covariance between $N_{bi}$ and
$\hat{\delta}_b (\mathbf{x}).$ In practice, $\hat{V}$ is biased
upwards, especially for small or moderate $B$. A bias-corrected
version is given by
\begin{equation} \label{SE-corrected-0}
\hat{V}_{c} = \hat{V} - \frac{1}{B^2} \sum_{i=1}^n \sum_{b=1}^B (
Z_{bi} - \bar{Z}_i )^2.
\end{equation}
Further assuming approximate independence of $N_{bi}$ and
$\hat{\delta}_b(\mathbf{x})$, another bias-corrected version is
given by
\begin{equation} \label{SE-corrected}
\hat{V}_{c} = \hat{V} - \frac{n-1}{B^2} \sum_{b=1}^B \{
\hat{\delta}_b(\x) - \hat{\delta}(\x)\}^2,
\end{equation}
which is easier to compute than (\ref{SE-corrected-0}).
\end{proposition}
The validity of these SE formulas will be investigated by
simulation in Section \ref{sec-simulation}. The bias-corrected SE
formulas in (\ref{SE-corrected-0}) and (\ref{SE-corrected})
generally yield very similar results with superior performance to
the uncorrected version (\ref{SE-IJ}). Note that computing
(\ref{SE-corrected-0}) entails evaluation of the matrix
$\mathbf{Z}= \left( Z_{bi} \right)$ at each different
$\mathbf{x}$. Therefore, the SE given in (\ref{SE-corrected}) is
recommended for its enhanced computational efficiency.

\subsection{Comparison with SR}
\label{sec-MSE}

Under the potential outcome framework, separate regression (SR) is
conventionally used to estimate $\delta(\mathbf{x})$; see, e.g.,
\citet{van der Laan:2006} and \citet{Foster:2011}. The basic idea
is to build a model based on data for treated individuals only to
estimate $\mu_1(\mathbf{x})=E(Y_1 \, | \, \mathbf{X} =
\mathbf{x})$ and build a model based on data for untreated
individuals only to estimate $\mu_0(\mathbf{x})=E(Y_0\, |\,
\mathbf{X}= \mathbf{x}).$ Let $\hat{\mu}_0(\mathbf{x})$ and
$\hat{\mu}_1(\mathbf{x})$ denote the resultant estimates of
$\mu_0(\mathbf{x})$ and $\mu_1(\mathbf{x})$, respectively. Then
ITE can be estimated as
\begin{equation}
\tilde{\delta}(\mathbf{x}) = \hat{\mu}_1(\mathbf{x}) -
\hat{\mu}_0(\mathbf{x}). \label{ITE-SR}
\end{equation}
Since SR essentially involves predictive modeling, random forests
\citep{Breiman:2001} are commonly used in the literature.

We would like to argue that RFIT is superior to SR. This is primarily because RFIT works on a simpler problem. To explain, consider the model form $Y = \mu_0(\x) + T \delta(\x) + \varepsilon,$ where $\mu_1(\x) = \mu_0(\x) + \delta(\x).$ Functions $\mu_0(\x)$ and $\delta(\x)$ may involve different sets of covariates. In the clinical setting, covariates showing up in $\mu_0(\x)$ only are called prognostic factors while covariates showing up in $\delta(\x)$ are called predictive factors (see, e.g., \citeauthor{Ballman:2015}, \citeyear{Ballman:2015}). In other words, predictive factors interact with the treatment and hence cause differential treatment effects. In SR, both $\mu_1(\x)$ and $\mu_0(\x)$ have to be estimated to have the difference $\delta(\x)$; thus it has to take both prognostic and predictive factors into consideration. Comparatively,  RFIT estimates $\delta(\x)$ directly by focusing on predictive factors only. This is because a prognostic factor won't cause a difference in differences, referring to its splitting statistic in (\ref{Q-GS}). In the following, we introduce a performance measure for RFIT and SE in estimating ITE $\delta(\x)$ and a theoretical understanding of the measure is attempted.

Both RFIT and SR take the bootstrap-based ensemble learning
approach; the ITE estimates $\hat{\delta}(\mathbf{x})$ in
(\ref{ITE-RFIT}) and $\tilde{\delta}(\mathbf{x})$ in
(\ref{ITE-SR}) involve randomness owing to bootstrap resampling,
the current data $\mathcal{D}$, and the point $\mathbf{x}$ at
which the estimation is made. To compare RFIT with SR, we consider
an average mean squares error (AMSE) measure defined by
\begin{equation}
\mbox{AMSE} = E_{\mathbf{X}, \mathcal{D}, \mathcal{B}}
\{\hat{\delta}(\mathbf{X}) - \delta(\mathbf{X}) \}^2, \label{AMSE}
\end{equation}
where the expectation is taken with respect to the bootstrap
distribution $\mathcal{B}$ given the current data $\mathcal{D}$,
the sampling distribution of data $\mathcal{D}$, and then the
distribution of $\mathbf{X}.$

Define
\begin{equation}
\bar{\delta}(\mathbf{x}; \mathcal{D}) = E_{\mathcal{B}} \,
\{\hat{\delta}(\mathbf{x})\},  \mbox{~~~and~~~}
\bar{\delta}(\mathbf{x}) = E_{\mathcal{D}} \,
\{\bar{\delta}(\mathbf{x}; \mathcal{D})\}, \label{delta-bar}
\end{equation}
where $\bar{\delta}(\mathbf{x}; \mathcal{D})$ is the RFIT estimate
of $\delta(\mathbf{x})$ obtained with perfect bootstrap or $ B
\rightarrow \infty$ and $\bar{\delta}(\mathbf{x})$ is the perfect
bootstrap RFIT estimate if, furthermore, we are allowed to
recollect data $\mathcal{D}$ freely. Similarly, we define
$\{\bar{\mu}_0(\mathbf{x}; \mathcal{D}),
\bar{\mu}_0(\mathbf{x})\}$ on the basis of
$\hat{\mu}_0(\mathbf{x})$ and $\{\bar{\mu}_1(\mathbf{x};
\mathcal{D}), \bar{\mu}_1(\mathbf{x})\}$ on the basis of
$\hat{\mu}_1(\mathbf{x})$ in SR. In addition, define
\begin{equation}
\bar{\mu}_0 = E_{\mathbf{X}} \, \{\bar{\mu}_0(\mathbf{x})\}
\mbox{~~~and~~~} \mu_0 =  E_{\mathbf{X}} \{\mu_0(\mathbf{X})\} =
E(Y_0), \label{mu-bar}
\end{equation}
and similarly $\{\bar{\mu}_1, \mu_1\}.$ Proposition
\ref{prop-AMSE} provides a decomposition of AMSE for the ITE
estimate $\hat{\delta}(\mathbf{x})$ by RFIT and for
$\tilde{\delta}(\mathbf{x})$ by SR.
\begin{proposition} \label{prop-AMSE}
For the RFIT estimate $\hat{\delta}(\mathbf{x})$ in
(\ref{ITE-RFIT}),
\begin{equation}
\mbox{AMSE} = E_{\mathbf{X}, \mathcal{D}, \mathcal{B}} \left\{
\hat{\delta}(\mathbf{X}) - \bar{\delta}(\mathbf{X}; \mathcal{D})
\right\}^2 + E_{\mathbf{X}, \mathcal{D}}
\left\{\bar{\delta}(\mathbf{X}; \mathcal{D}) -
\bar{\delta}(\mathbf{X}) \right\}^2 + E_{\mathbf{X}} \left\{
\bar{\delta}(\mathbf{X}) - \delta (\mathbf{X}) \right\}^2.
\label{AMSE-RFIT}
\end{equation}
For the SR estimate $\tilde{\delta}(\mathbf{x})$ in
(\ref{ITE-SR}),
\begin{eqnarray}
\mbox{AMSE} &=& E_{\mathbf{X}, \mathcal{D}, \mathcal{B}} \left\{
\hat{\mu}_1(\mathbf{X}) - \bar{\mu}_1(\mathbf{X}; \mathcal{D})
\right\}^2 + E_{\mathbf{X}, \mathcal{D}}
\left\{\bar{\mu}_1(\mathbf{X}; \mathcal{D}) -
\bar{\mu}_1(\mathbf{X}) \right\}^2 + E_{\mathbf{X}} \left\{
\bar{\mu}_1(\mathbf{X}) - \mu_1 (\mathbf{X}) \right\}^2  \nonumber \\
 & + &  E_{\mathbf{X}, \mathcal{D}, \mathcal{B}} \left\{
\hat{\mu}_0(\mathbf{X}) - \bar{\mu}_0(\mathbf{X}; \mathcal{D})
\right\}^2 + E_{\mathbf{X}, \mathcal{D}}
\left\{\bar{\mu}_0(\mathbf{X}; \mathcal{D}) -
\bar{\mu}_0(\mathbf{X}) \right\}^2 + E_{\mathbf{X}} \left\{
\bar{\mu}_0(\mathbf{X}) - \mu_0 (\mathbf{X}) \right\}^2  \nonumber \\
 & - &  2 E_{\mathbf{X}} \left[\{\bar{\mu}_1(\mathbf{X}) - \mu_1(\mathbf{X})\}
\{\bar{\mu}_0(\mathbf{X}) - \mu_0(\mathbf{X})\} \right].
\label{AMSE-SR}
\end{eqnarray}
\end{proposition}
The first term of AMSE in (\ref{AMSE-RFIT}) corresponds to Monte
Carlo variation resulted from using a finite number of $B$
bootstrap samples. The second term represents the sampling
variation owing to lack of endless supply of training data in
reality. The third term is the bias. Similar interpretation holds
true for the terms in (\ref{AMSE-SR}), yet with an additional
covariance term $-2 E_{\mathbf{X}} \left[\{\bar{\mu}_1(\mathbf{X})
- \mu_1(\mathbf{X})\} \{\bar{\mu}_0(\mathbf{X}) -
\mu_0(\mathbf{X})\} \right].$

Ensemble learners such as RF and bagging aim for variance
reduction by imitating the endless supply of replicate data via
bootstrap resampling. This is why we have the additional
decomposition $$ E_{\mathbf{X}, \mathcal{D}, \mathcal{B}} \{
\hat{\delta}(\mathbf{X}) - \bar{\delta}(\mathbf{X}) \}^2 =
E_{\mathbf{X}, \mathcal{D}, \mathcal{B}} \left\{
\hat{\delta}(\mathbf{X}) - \bar{\delta}(\mathbf{X}; \mathcal{D})
\right\}^2 + E_{\mathbf{X}, \mathcal{D}}
\left\{\bar{\delta}(\mathbf{X}; \mathcal{D}) -
\bar{\delta}(\mathbf{X}) \right\}^2$$ in (\ref{AMSE-RFIT});
similarly for $\hat{\mu}_1(\mathbf{X})$ and
$\hat{\mu}_0(\mathbf{X})$ in (\ref{AMSE-SR}). However, ensemble
learning has little effect on the bias term $E_{\mathbf{X}}
\{\bar{\delta}(\mathbf{X}) - \delta(\mathbf{X}) \}^2$ in
(\ref{AMSE-RFIT}); similarly for the two bias terms in
(\ref{AMSE-SR}) as well as the covariance term $-2 E_{\mathbf{X}}
\left[\{\bar{\mu}_1(\mathbf{X}) - \mu_1(\mathbf{X})\}
\{\bar{\mu}_0(\mathbf{X}) - \mu_0(\mathbf{X})\} \right].$ The bias
problem for ensemble learners such as random forests has been
noted by \citet{Breiman:1999} and others. From another
perspective, RF facilitates a smoothing procedure by averaging
data over an adaptive neighborhood; as a result, it cuts the hill
and fills the valley.

While both RFIT and SR would suffer from certain bias, the AMSE in
SR tends to be larger than that of RFIT in general as we shall demonstrate numerically in Section \ref{sec-simulation}. Numerical evidence shows that SR is more prone to the bias problem because it tends to underestimate a large ITE and overestimate a small ITE. In fact, such a bias also has an effect on the last covariance term in (\ref{AMSE-SR}). A large ITE
$\delta(\mathbf{x})$ occurs when $\mu_1(\mathbf{x})$ is large and/or
$\mu_0(\mathbf{x})$ is small. The smoothing effect yields $\bar{\mu}_1(\mathbf{X}) - \mu_1(\mathbf{x}) <0$ with cut hills
and $\bar{\mu}_0(\mathbf{X}) - \mu_0(\mathbf{X}) >0$ with filled valleys. Thus $\{\bar{\mu}_1(\mathbf{X}) - \mu_1(\mathbf{X})\} \{\bar{\mu}_0(\mathbf{X}) - \mu_0(\mathbf{X})\}$ tends to be negative. A similar observation holds for a small ITE, which occurs when $\mu_1(\mathbf{x})$ is small and/or $\mu_0(\mathbf{x})$ is large. As a result, the last term in
(\ref{AMSE-SR}) tends to be negative, leading to a more inflated AMSE for SR.

\section{Simulation Studies}
\label{sec-simulation} This section presents  results from
simulation studies designed to compare the smooth sigmoid
surrogate (SSS) splitting method with greedy search (GS) in
finding the best cutoff point; compare RFIT with separate
regression (SR) in estimating the individualized treatment effects
(ITE); and investigate the standard error (SE) formulas for the
estimated ITE.

\subsection{Comparison of SSS versus GS}
\label{sec-simulation-SSS-GS} To compare SSS with GS, we generated
data from model
\begin{equation}
 y = 0.5 + 0.5 \, T + 0.5 \, \Delta + 0.5 \cdot
T\, \Delta + \varepsilon, \label{Model-A}
\end{equation}
where $\Delta = \Delta(x; c_0) = I(x \geq c_0)$, $x \sim
\mbox{uniform}[0, 1],$ $c_0=0.5$, and $\varepsilon \sim N(0, 1).$ We considertwo sample sizes $n=50$ and $n=500$, corresponding to relatively
fewer and larger numbers of observations in a node. For each
simulated data, both GS and SSS are used to identify the best
cutoff point $\hat{c}.$ Different $a=1, 2, \ldots, 100$ values are
used in SSS. For each model configuration, 500 simulation runs are
made.

Figure \ref{fig02} presents the empirical density and the MSE
measure, defined as $\mbox{MSE} = \sum_{k=1}^{500} (\hat{c}_m -
c_0)^2/500$, for the estimated cutoff point $\hat{c}$ by SSS and
GS. It can be seen that SSS compares favorably to GS in terms of
MSE. From the empirical density plots, it can be seen that one
important contribution made by SSS is the reduced variation as
compared to GS. In most scenarios, SSS shows considerable
stability with respect to the choice of the shape parameter $a$,
especially with relatively larger $a$ values. Too small an
$a \leq 5$ value can result in deteriorated performance and hence is not advisable. It seems desirable to balance a more accurate approximation to the indicator function with a relatively larger $a$ value and the more smooth objective function for optimization with a relatively smaller $a$ value. We would like to comment that estimating $a$ is not a good idea in the tree setting for several reasons: tree models seek threshold effects which entail a relatively large $a$ value; estimating $a$ unnecessarily slows down the computation at each node split; having a different $a$ for each covariate will make the results less comparable across covariates in order to find the best split. Recall that SSS works with standardized $X$ and transfers $\hat{c}$ back to its original scale; we recommend fixing $a=10$ in SSS for standardized $x$ based on our more extensive numerical
explorations that are not presented here. Henceforth, SSS with
$a=10$ is used by default in the RFIT implementation.

To compare the computing time, we generated data from the same
model; a slight modification was made in the way of simulating the
covariate: $x$ follows a discrete uniform distribution over
$\{1/K, 2/K, \ldots, K/K\}$ so that $x$ has a total of $K$
distinct values. This allows us to investigate the computing time
with different $K$. The choices for $n$ and $K$ are $n \in \{50,
100, 500, 1000, 2000, 10000\}$ and $K \in \{10, 100, 500\}.$
Table \ref{tbl01} tabulates the computing time in seconds for GS
and SSS splitting, averaged over 10 simulation runs for each
setting. It can be seen that SSS is superior to GS in terms of
computational efficiency. As expected, it takes longer for both GS
and SSS as $n$ increases in general. It takes longer for GS as $K$
increases; but this is not the case for SSS.

\subsection{Comparison of RFIT versus SR}
\label{sec-simulation-RFIT-SR} We compare SSS to SR in estimating
ITE. The data are generated with the following scheme: first simulate
five ($p=5$) predictors $x_j \sim \mbox{uniform}[0, 1]$ for $j=1,
\ldots, 5$; then we generate $y'_0 = \mu_0(\mathbf{x}) + \alpha +
\varepsilon_0$ with a nonlinear polynomial
$$ \mu_0(\mathbf{x}) = -2 - 2 x_1 - 2 x_2^2 + 2 x_3^3$$
and $\alpha$ and $\varepsilon_0$ being independent from
$\mathcal{N}(0,1);$ next, we generate $y'_1 = \mu_1(\mathbf{x}) +
\alpha + \varepsilon_1,$ where $\mu_1(\mathbf{x}) =
\mu_0(\mathbf{x}) + \delta(\mathbf{x})$ and $\varepsilon_1 \sim
\mathcal{N}(0,1)$ is independent of both $\alpha$ and
$\varepsilon_0$. A random effect term $\alpha$ is introduced to
mimic some common characteristics shared by repeated measures
$Y'_0$ and $Y'_1$ taken from the same subject. The unit-level
effect $Y'_1 - Y'_0$ equals $\delta(\mathbf{x}) + (\varepsilon_1 -
\varepsilon_0)$, where $(\varepsilon_1 - \varepsilon_0)$
represents additional random errors that can not be accounted for
by covaraites $\mathbf{x}.$ Four models (I)--(IV) are considered
for the ITE $\delta(\mathbf{x})$, as given below:
\begin{eqnarray}
& \mbox{Model I:~~} & \delta(\mathbf{x}) =  -2 + 2 x_1 + 2 x_2 \label{Model-I} \\
& \mbox{Model II:~} & \delta(\mathbf{x}) =  -2 + 2 \, I(x_1 \leq 0.5) + 2 \, I(x_2 \leq 0.5) \, I(x_3 \leq 0.5) \label{Model-II} \\
& \mbox{Model III:} & \delta(\mathbf{x}) =  -6 + 0.1 \exp(4x_1) + 4 \exp\{20(x_2 - 0.5)\} + 3 x_3 + 2 x_4 + x_5 \label{Model-III}\\
& \mbox{Model IV:} & \delta(\mathbf{x}) =  -10 + 10 \sin (\pi x_1
x_2) + 20 (x_3 - 0.5)^2 + 10x_4 + 5 x_5. \label{Model-IV}
\end{eqnarray}
Model I exemplifies a linear ITE; Model II represents a
tree-structured model; Model III \& IV are derived from two
nonlinear models in \citet{Friedman.1991}. Finally, we simulate
the randomized treatment assignment variable $T$ independently
from Bernoulli(0.5) and hence the observed response $y = T y'_1 +
(1-T) y'_0.$

For each training data set $\mathcal{D}$, both RFIT and SR are
used to learn a model on ITE. In order to evaluate their
performance, a test sample $\mathcal{D}'$ of size $n' = 2000$ is
generated beforehand. The ITE models trained with RFIT and SR in each simulation are applied to estimate the ITE for $\mathcal{D}'$ and a mean square error (MSE) measure $\mbox{MSE} = \sum_{i=1}^{n'} \{\hat{\delta}(\mathbf{x}_i) - \delta(\mathbf{x}_i) \}^2/n'$ is computed. Two sample sizes $n=100$ and $n=500$ are considered for the training data
$\mathcal{D}$ and a total of 200 simulation runs is used for each
simulation setting.

Figure \ref{fig03} plots the parallel boxes of MSE measures from
200 simulation runs for RFIT and SR. The averages are highlighted
with blue bars, corresponding to estimates of the AMSE in
(\ref{AMSE}). It can be seen that RFIT outperforms SR consistently
in all the scenarios considered here. Again, the superiority of RFIT can be explained by the fact that it works on an easier task than SR by estimating $\delta(\x)$ directly. Additional numerical insight into the bias problem is provided by plotting the estimated ITE $\hat{\delta}(\mathbf{x})$ (averaged over 200 simulated runs) versus the actual ITE $\delta
(\mathbf{x})$. See Section \ref{sec-SM-numerical} of the
Supplementary Material.

\subsection{Standard Error Formulas} \label{sec-simulation-SE}

To investigate the validity and performance of the standard error
(SE) formulas for estimated ITE, we generated training data sets
of size $n=500$ from Model III in (\ref{Model-III}) and one test
data set $\mathcal{D}'$ of size $n'=50.$ For each training data
set $\mathcal{D}$, $B=2000$ bootstrap samples is used to train
RFIT and then the trained RFIT is applied to estimate ITE for each
observation in $\mathcal{D}'$ together with standard errors. We
repeat the experiment for 200 simulation runs. At the end of the
experiment, we have 200 predicted ITE $\hat{\delta}$ for each
observation in $\mathcal{D}'$, together with 200 SEs. Accordingly,
we compute the standard deviation (SD) of these  ITE estimates
$\hat{\delta}$ and average the SE values. If the SE formula works
well, the SE values should be close to their corresponding SD
values.

Figure \ref{fig04} plots the averaged SE versus SD for each
observation in the test sample $\mathcal{D}'$. It can be seen that
the uncorrected standard errors are overly conservative. After the
bias-correction, they become reasonably close to the SD values.
The bias-corrected SE presented here is computed from
(\ref{SE-corrected}). The other version (\ref{SE-corrected-0})
that is somewhat harder to compute provides very similar results,
which have been omitted from the plotting.

We experimented with other models in Section
\ref{sec-simulation-RFIT-SR} and similar results were obtained.
One issue pertains to the number $B$ of bootstrap samples needed.
According to \citet{Efron:2014}, a large $B$, e.g., $B=2,000$ is
needed to guarantee the validity of IJ-based standard errors. We
experimented with different $B$ values. Generally speaking, ITE
estimation stabilizes quickly even with a small $B$, e.g.,
$B=100$; however, negative values may frequently occur to the
bias-corrected variance estimates in both (\ref{SE-corrected-0})
and (\ref{SE-corrected}) when $B$ is small or moderate, e.g.,
$B=500.$ Thus a large number $B$ of bootstrap samples are needed
to have sensible results for the SE formulas.

\section{Application: Acupuncture Trial}
\label{sec-example}

For further illustration of RFIT, we consider data collected from
a acupuncture headache trial \citep{Vickers:2004}, available at
\newline
\centerline{\url{https://trialsjournal.biomedcentral.com/articles/10.1186/1745-6215-7-15}}
In this randomized study, 401 patients with chronic headache,
predominantly migraine, were randomly assigned either to receive
up to 12 acupuncture treatments over three months or to a control
intervention offering usual care. Among many other measurements,
the primary end point of the trial is the change in headache
severity score from baseline to 12 months since study entry. The
acupuncture treatment was concluded effective overall by
significantly bringing down the headache score and other outcome
measures. More details of the trial and the results are reported
in \citet{Vickers:2004}.

To apply RFIT, we consider only the 301 participants who completed
the trial. The response variable is taken as the difference in
headache severity score between baseline and 12 months, while the
score at baseline is treated as a covariate. A total of 18
covariates are included in the analysis, which are essentially
demographic, medical, or treatment variables measured at baseline.
See Table \ref{tblS1} in the Supplementary Materials for a brief
variable description.

A total of $B=2,000$ trees are used to build RFIT, where the scale
parameter $a$ is set as $a=10$ in SSS splitting. ITE  is estimated
for each individual in the same data set and the IJ-based standard
error (SE) with bias correction is also computed. Figure
\ref{fig05} provides a bar plot of the estimated ITE, plus and
minus one SE, sorted by ITE. It can be seen that a majority of ITE are above 0, indicating the effectiveness of acupuncture. Overall speaking, the treatment effects in this trial show certain heterogeneity, but
not by much. It is interesting to note that the averaged ITE is
3.9. Comparatively,  the unadjusted mean difference in headache
score is 6.5 while the adjusted effect from ANCOVA is 4.6, as
reported in Table 2 of \citet{Vickers:2004}. Figure \ref{fig05}
also shows many individuals, for whom the acupuncture treatment
did not help much, including two individuals, the 44-th  (with
patient ID 222) and the 224th (with patient ID 630). Both are
female patients aged 60 and 58, suffer migraine, and were assigned
to the control group, but surprisingly achieved a reduction of 36
and 29.75 in severity score, respectively. Their initial severity
scores are also relatively similar: 44.25 and 37. Their estimated
ITEs turn out to be $-14.81$ and $-9.09$, indicating a detrimental
effect from acupuncture. Although the performances of these two
patients are quite unusual relative the the remainder of the
patients, they may indicate a small subgroup that is worth further
investigation.

\iffalse
One important feature of random forests is variable importance
ranking. \citet{Su:2009} extended the technique to interaction
trees, with an aim to seek important effect moderators or
modifiers. We modified this method by incorporating the SSS
splitting. Figure \ref{fig05}(b) plots the resultant variable
importance values for the acupuncture headache data. It can be
seen that the baseline severity score \texttt{pk1} is the leading
effect modifier for acupuncture, followed by general health
\texttt{gen1}, headache frequency \texttt{f1}. With some
additional exploration, it can be found that these important
effect moderators highlight larger effects on patients with more
severe symptoms and worse health conditions at baseline.
\fi

\section{Discussion}
\label{sec-discussion}

We have tackled the problem of estimating individualized treatment
effect (ITE) by using the random forests of interaction trees
(RFIT). Smooth sigmoid surrogate (SSS) splitting is introduced to
speed up RFIT and possibly improve its performance. We have also
applied the infinitesimal jackknife method to derive a standard
error for the estimated ITE. Altogether, RFIT provides enlightening results in deploying personalized medicine by informing a new patient about the potential efficacy of the treatment on him/her.

According to our numerical experiments, RFIT outperforms the
commonly used separate regression (SR) approach for estimating
ITE. SR estimates the potential outcomes separately and then takes
difference. In RFIT, however, we first group individuals so that
those with similar treatment effects are put together and then
estimate the treatment effect by taking differences within each
group. Comparatively, RFIT focuses on predictive covariates and estimation of ITE directly while SR has to deal with prognostic covariates and works on a harder problem. Since SR has been widely used as an intermediary step in other causal inference procedures, our method might contribute to their improvement as well.

To conclude, we identify several avenues for future research.
First of all, our discussion has been restricted to data from
randomized experiments. Assessing treatment effects with data from
observational data can be very different, entailing adjustment for
potential confounders. See, e.g., \citet{Su:2012} and
\citet{Wager:2016}. Secondly, the standard error formula provides
some assessment for precision in estimating ITE; however, issues
such as consistency of RFIT, asymptotic normality of estimated ITE (see comments in \citeauthor{Efron:2014}, \citeyear{Efron:2014}) and multiplicity have not been thoroughly addressed as of yet. Thirdly, besides variable importance ranking, several other useful features from
random forests including partial dependence plots and proximity
matrix \citep{Liaw:2002} have yet to be explored for RFIT.

\section*{Acknowledgements}

XS was partially supported by NIMHD grant 2G12MD007592 from NIH;
LL was partially supported by AHRQ grant HS 020263; RL was
supported in part by NSF grant 1633130.

\iffalse
\vspace{0.2in}
\appendix
\begin{center}
{\Large APPENDIX}
\end{center}

\section{Proofs}
\fi

%\bibliographystyle{plainnat}

% TABLE AND FIGURES

\newpage

\renewcommand{\tabcolsep}{8.pt}
\renewcommand{\arraystretch}{1.1}
\renewcommand{\baselinestretch}{1.1}
\begin{table}[h]
\caption{Computing time comparison between smooth sigmoid
surrogate (SSS) and greedy search (GS) in finding the best cutoff
point for one covariate. Entries are the computing times (in
seconds) averaged over 10 runs.} \vspace{.4in}
% \rule{5.5in}{.03cm}
\centering
\begin{tabular}{rcccccccc} \hline \hline
    &   \multicolumn{2}{c}{$K=10$}          &&  \multicolumn{2}{c}{$K=100$}         &&  \multicolumn{2}{c}{$K=500$}         \\ \cline{2-3} \cline{5-6} \cline{8-9}
    &   GS  &   SSS &&  GS  &   SSS &&  GS  &   SSS \\ \hline
$n=~~~50$   &   0.000   &   0.001   &&  0.003   &   0.000   &&  0.003   &   0.000   \\
100 &   0.000   &   0.000   &&  0.006   &   0.000   &&  0.003   &   0.004   \\
500 &   0.002   &   0.000   &&  0.012   &   0.003   &&  0.047   &   0.000   \\
1000    &   0.004   &   0.000   &&  0.023   &   0.002   &&  0.100   &   0.000   \\
2000    &   0.003   &   0.004   &&  0.038   &   0.005   &&  0.201   &   0.003   \\
5000    &   0.008   &   0.002   &&  0.094   &   0.002   &&  0.462   &   0.001   \\
10,000  &   0.017   &   0.005   &&  0.182   &   0.005   &&  0.899   &   0.010   \\ \hline
\end{tabular}
% \rule{8.2in}{.03cm}
\label{tbl01}
\end{table}

\renewcommand{\baselinestretch}{1.0}
\begin{figure}[h]
\centering
  \includegraphics[scale=0.54, angle=0]{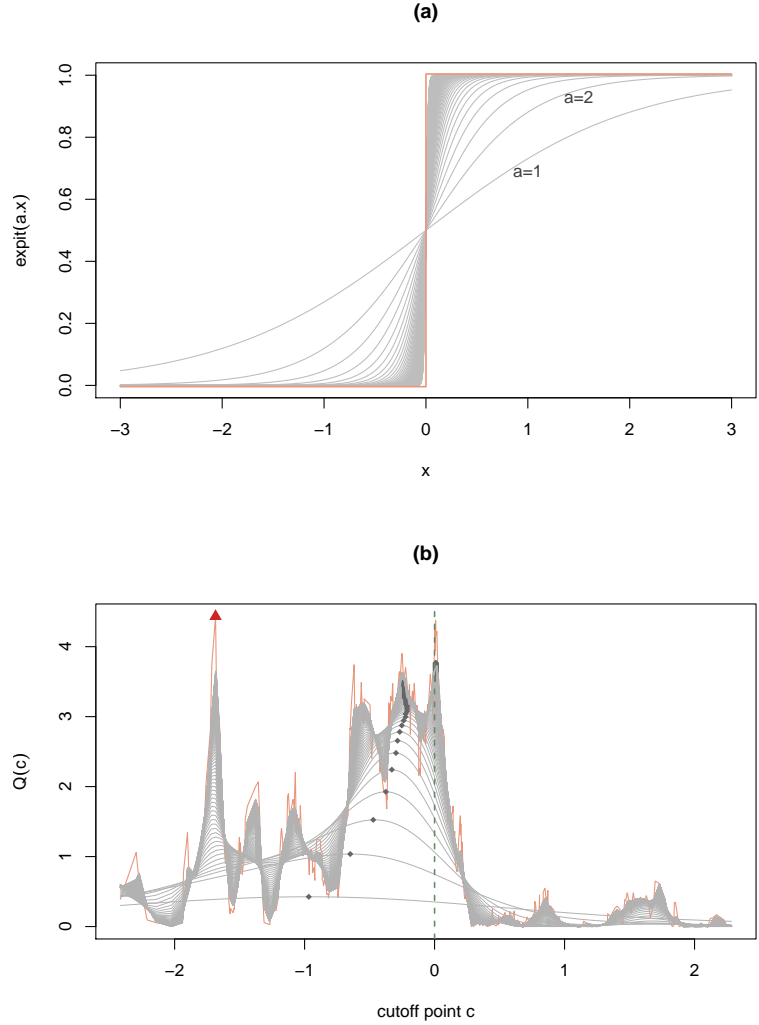}
  \caption{Illustration of Smooth Sigmoid Surrogate (SSS)
for Splitting Data: (a) The discrete threshold function $\Delta(x;
c) = I(x \geq c)$ with $c=0$ (in {\color{orange}{orange}}) and its
expit approximation  $s(x; c) = \expit\{ a \, x-c) \}$ (in
{\color{gray}{gray}}); (b) The splitting statistic $Q(c)$ computed
at each cutoff point $c$ in greedy search and its SSS
approximations with $a= \{1, 2, \ldots, 100 \}.$ In panel (b),
data of size $n=500$ are generated from model $y = 0.5 + 0.5 \, T
+ 0.5 \, \Delta + 0.5 \cdot T\, \Delta + \varepsilon$, where
$\Delta = \Delta(x; c_0)$ with true cutoff point $c_0=0$
(indicated by the {\color{green}{green}} dashed vertical line) and
both $x$ and $\varepsilon$ are from $N(0, 1).$ The best cutoff
found by GS is denoted by the {\color{red}{red}} triangle while
the {\color{black}{black}} diamond dots indicate the best cutoff
points found by SSS with different $a$ values. \label{fig01}}
\end{figure}

\begin{figure}[h]
\centering
  \includegraphics[scale=0.57, angle=270]{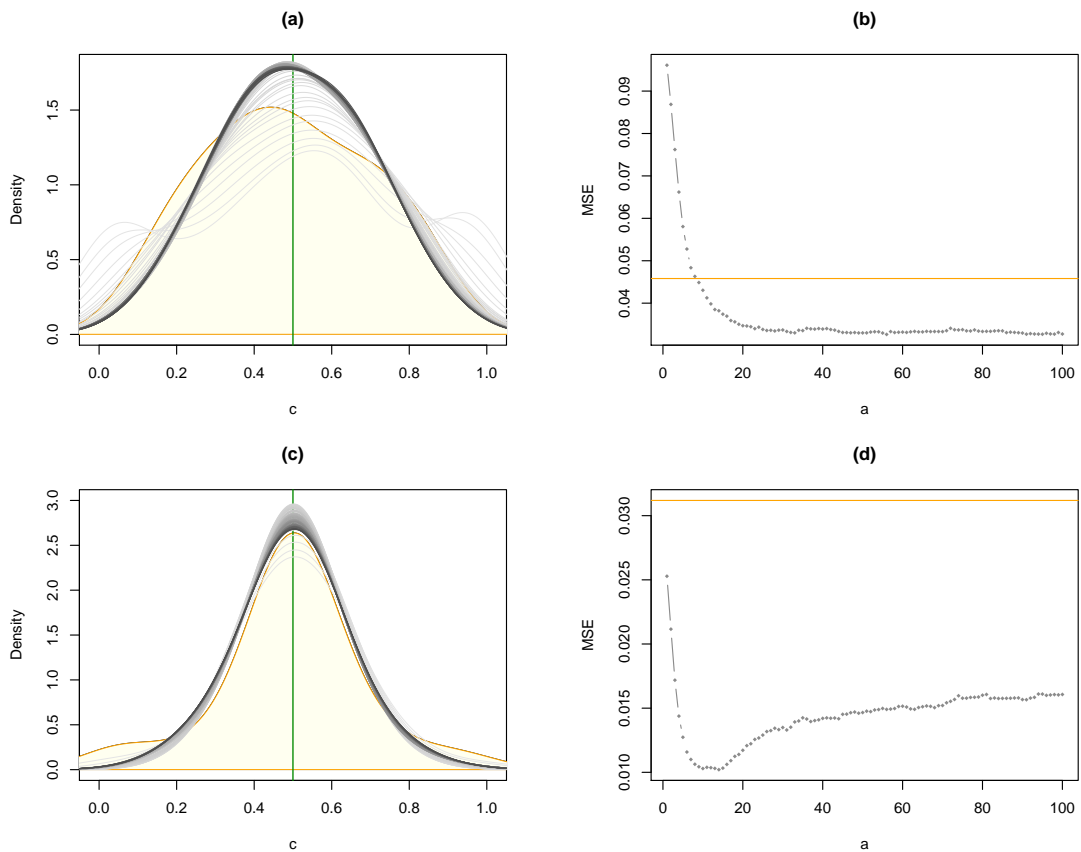}
  \caption{Comparing SSS with GS in Finding the Best Cutoff Point:
the left panels present the empirical density of $\hat{c}$ from
500 runs found by GS (shaded in {\color{orange}{orange}}) and SSS
(in {\color{gray}{grayscale}}) with $a=1, 2, \ldots , 100$, where
the true cutoff point $c_0=0.5$ is indicated by the
{\color{green}{green}} vertical bar; the right panels present MSE
measures of SSS for different $a$ values, where the horizonal
{\color{orange}{orange}} line corresponds to the MSE from GS.
Panels differ in terms of sample size $n$: $n=50$ in Panels (a) \&
(b); $n=500$ in Panels (c) \& (d). \label{fig02}}
\end{figure}

\begin{figure}[h]
\centering
  \includegraphics[scale=0.65, angle=270]{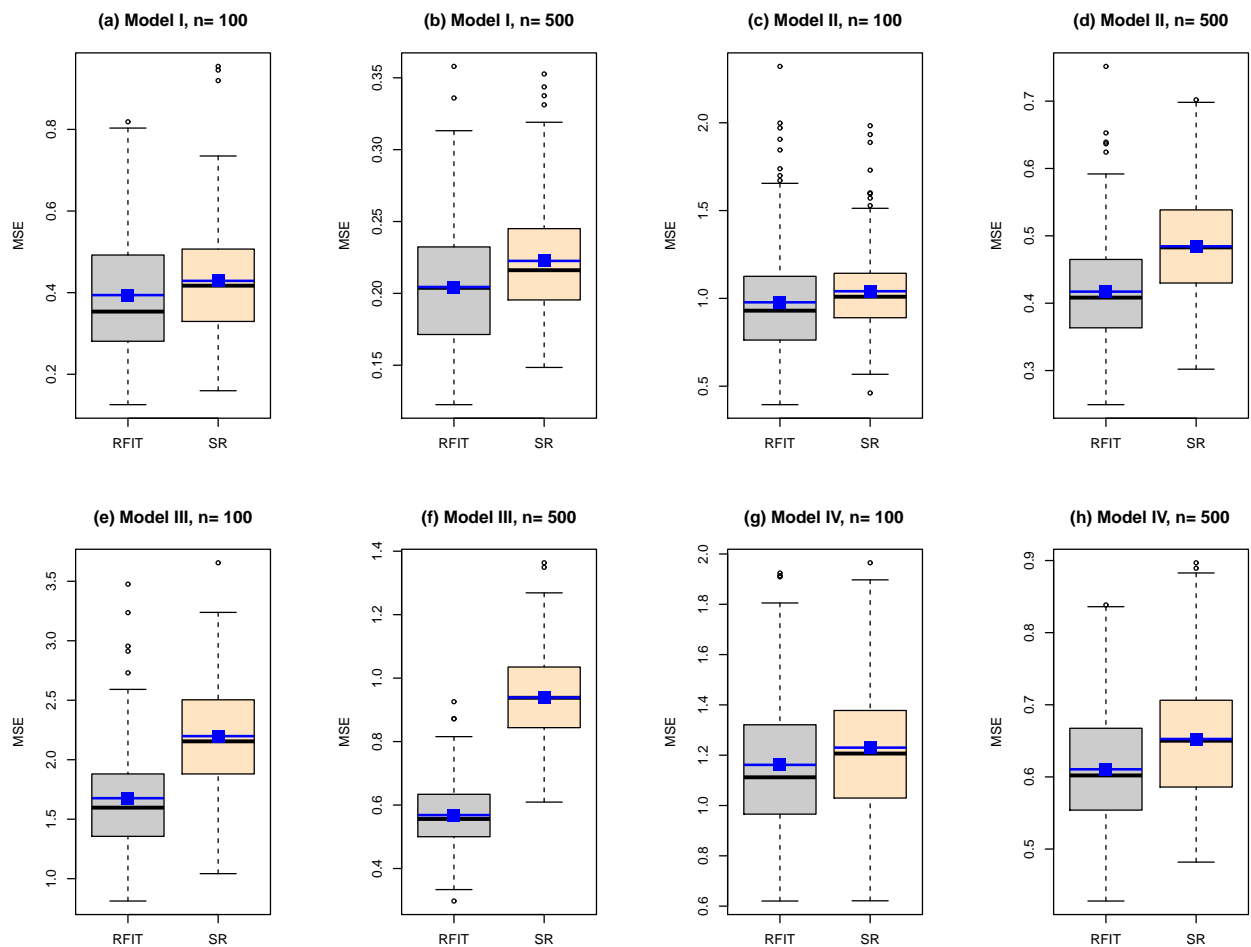}
  \caption{Comparing RFIT with Separate Regression (SR) in Estimating
ITE: parallel boxplots of MSE values are based on a test sample of
$n'=2000$ with 200 simulation runs. The {\color{blue}{blue}}
middle bar indicates the average of MSE measures. \label{fig03}}
\end{figure}

\begin{figure}[h]
\centering
  \includegraphics[scale=0.58, angle=0]{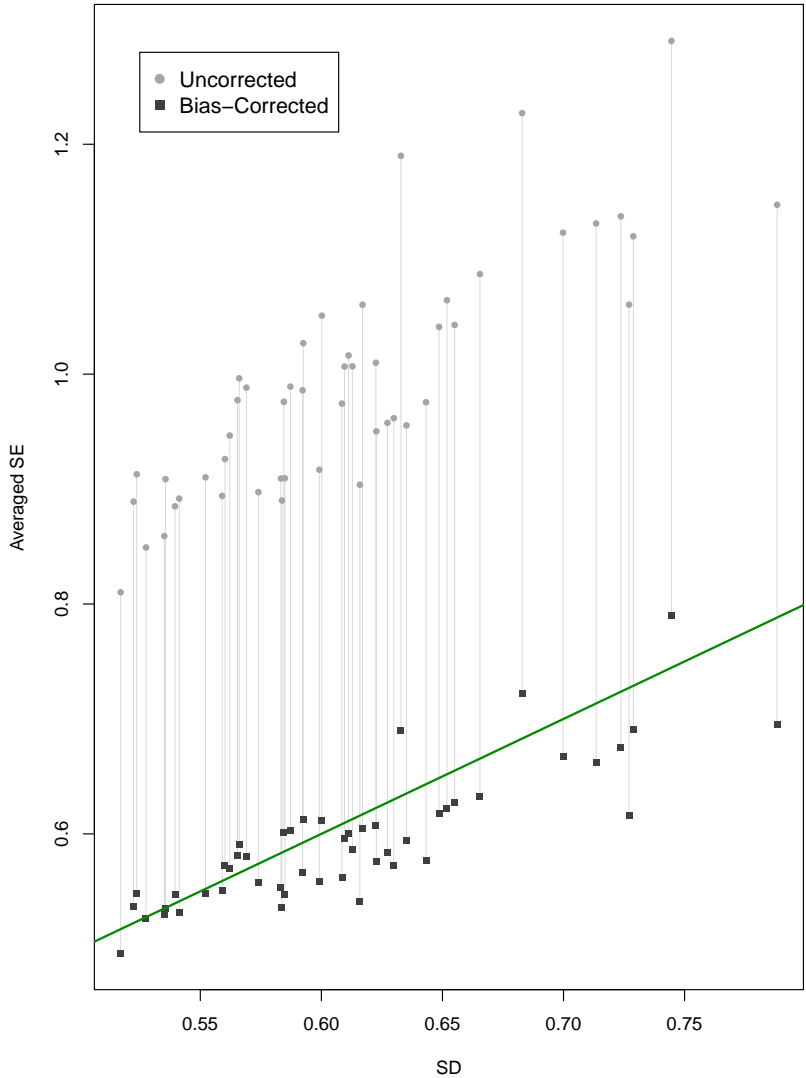}
  \caption{Plot of averaged standard errors (SE) versus sample standard
deviation (SD) of predicted ITE $\hat{\delta}(\mathbf{x})$ for
$n'=50$ observations in a test sample. The standard deviation (SD)
are computed based on 200 simulation runs while the standard
errors (SE) are averaged over the 200 runs. In each simulation
run, a training sample of size $n=500$ is generated from Model III
in (\ref{Model-III}) and a bootstrap size $B=2000$ is used to
build RFIT. The bias-corrected and uncorrected SE averages for the
same observation are connected with a {\color{gray}{gray}} line
segment. The reference line in {\color{green}{green}} is $y=x.$
\label{fig04}}
\end{figure}

\begin{figure}[h]
\centering
  \includegraphics[scale=0.65, angle=270]{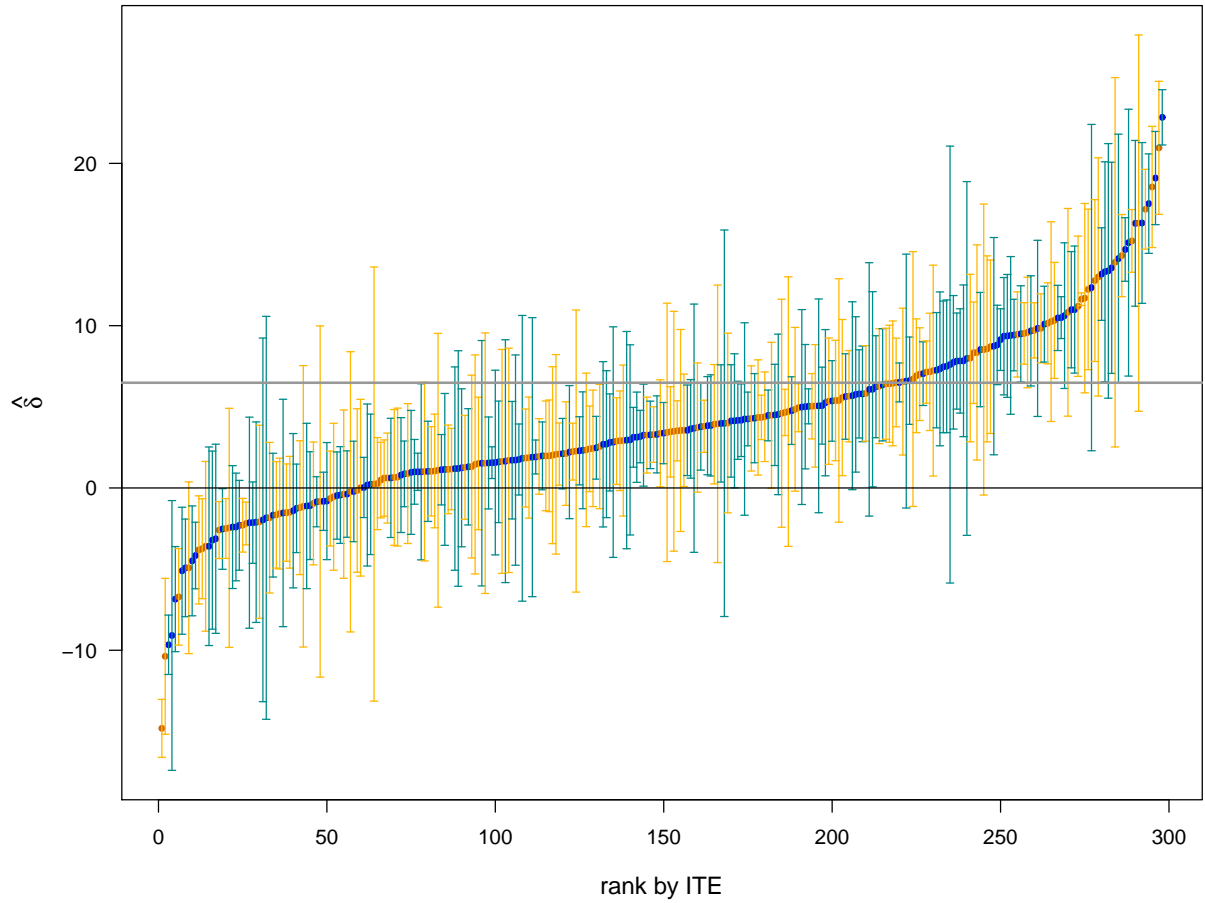}
  \caption{RFIT Analysis of the Headache Data: the error bar plot of the estimated ITE $\pm$
  SE. Individuals are ranked by estimated ITE. The {\color{gray}{gray}} horizontal
line indicates the unadjusted average treatment effect 6.484,
i.e., the mean difference in headache severity score.
\label{fig05}}
\end{figure}

%%%%%%%%%%%%%%%%%%%%%%%%%%%%%%%%%%%%%%%
%%%%%%%%%%%%%%%%%%%%%%%%%%%%%%%%%%%%%%%
%% Supplementary Materials
%%%%%%%%%%%%%%%%%%%%%%%%%%%%%%%%%%%%%%%
%%%%%%%%%%%%%%%%%%%%%%%%%%%%%%%%%%%%%%%

\renewcommand{\baselinestretch}{1.0}

\newpage
\clearpage \setcounter{page}{1} \setcounter{table}{0}
\setcounter{figure}{0}
% Numbering tables & figures A1, A2, etc
% \renewcommand{\thetable}{B\arabic{table}}
% \renewcommand{\thefigure}{B\Roman{figure}}
\renewcommand{\thetable}{\Roman{table}}
\renewcommand{\thefigure}{\Roman{figure}}
\numberwithin{equation}{section}

\begin{center}
{\large \textbf{Supplementary Materials to}} \\
{\Large \textbf{``Random Forests of Interaction Trees for Estimating \\
 Individualized Treatment Effects in Randomized
Trials"}}
\end{center}

\appendix
\section{Proofs}
\label{sec-proofs} This section contains proofs to the
propositions.

\subsection{Proof of Proposition
\ref{prop-computational-complexity}}
\label{sec-proof-prop-computing} When the GS splitting is
conducted without updating, GS evaluates the splitting criterion
$Q(c)$ for every distinct splitting points of $X$. For each $c$,
computation of $Q(c)$ is $O(n).$ Note that extracting the unique
values of $X$ is $O(n)$ in general. Therefore, the total
complexity amounts to $O(Kn)$ in this case, where $K$ is the
number of distinct values of $X.$ Since $X$ is continuous with
$K=O(n),$ the complexity becomes $O(n^2).$

GS can be alternatively done via an updating formula. This entails
sorting the $y_i$ and $T_i$ values in the ascending (or
descending) order of distinct $X$ values so that the $Q(c)$ value
at cutoff point $c$ can be conveniently obtained by utilizing its
previous value at the neighboring cutoff point. The computation
involved in the updating step itself is negligible and does not
escalate the computational complexity level $O(n)$ asymptotically.
Nevertheless, a stable algorithm for general-purpose sorting is at
best $O\{\ln(n) \, n\}.$ For example, the \texttt{sort} function
in R is either $O(n^{4/3})$ if Shellsort is used or $O\{\ln(n) \,
n\}$ if the Quicksort method is used.

For SSS, each iterative step in Brent's method involves evaluation
of $Q(c)$, which is $O(n).$ SSS requires standardization of $X$
and transformation of $\hat{c}$ back to the original scale, both
operations being $O(n)$. Put together, its complexity is $O(mn)$,
where $m$ is the number of iterative steps in the optimization
algorithm. \hfill $\blacksquare$

\subsection{Proof of Proposition \ref{prop-SE}}
\label{sec-proof-prop-SE} The proof essentially follows
\citet{Efron:2014} and \citet{Wager:2014}, with more details added
and some rewriting mainly for our own understanding. The arguments
are based on the the `ideal bootstrap' and treat the current data
$\mathcal{D}$ as fixed, where $B=n^n$ corresponds to all possible
choices when taking a bootstrap sample from $\mathcal{D}.$

Let $\mathcal{D}_b$ denote the $b$-th re-sample for $b=1, \ldots,
B.$ Introduce random vector $\mathbf{N}_{b} = (N_{b1}, \ldots,
N_{bn})^T,$ where $N_{bi}$ counts the frequency that the $i$-th
observation in $\mathcal{D}$ shows up in $\mathcal{D}_b$ with
$\sum_{i=1}^n \mathbf{N}_{b} = n.$ It follows that $\mathbf{N}_{b}
\sim \mbox{Multinomial}(n; \mathbf{p}_0)$ with $\mathbf{p}_0 =
(1/n, \ldots, 1/n)^T \in \mathbb{R}^n.$ Fixing $\mathcal{D}$,
$\mathbf{N}_b$ completely determines $\mathcal{D}_b.$ Thus, the
ITE estimate $\hat{\delta}_b(\mathbf{x})$ based on $\mathcal{D}_b$
can be written as a function $\hat{\delta}_b = T(\mathbf{N}_{b})$
of $\mathbf{N}_{b}.$ The RFIT estimate $\hat{\delta}(\mathbf{x})$
with ideal bootstrap is the expectation of $T(\mathbf{N}_{b})$,
i.e., $\hat{\delta}(\mathbf{x}) = E_{\mathbf{N}_b}
\{T(\mathbf{N}_{b})\}.$ Since the distribution of $\mathbf{N}_b$
is fully determined by the probability vector $\mathbf{p}_0$,
rewrite $\hat{\delta}(\mathbf{x})$ as a function
$\hat{\delta}(\mathbf{x}) = S(\mathbf{p})$ of $\mathbf{p}$ if more
generally $\mathbf{N}_{b} \sim \mbox{Multinomial}(n; \mathbf{p})$.
The symbol $S(\cdot)$ is used to denote this function, for
ensemble learners such as $\hat{\delta}(\mathbf{x})$ are
essentially obtained via `bootstrap smoothing'.

Define the influence function $\dot{S}_i$
\begin{equation}
\dot{S}_i  = \lim_{\epsilon \rightarrow 0^+} \frac{S(\mathbf{p}_0
+ \epsilon (\mathbf{e}_i - \mathbf{p}_0)) -
S(\mathbf{p}_0)}{\epsilon} \label{S-dot-i}
\end{equation}
for $i=1, 2, \ldots, n$, where $\mathbf{e}_i = (0, \ldots, 0, 1,
0, \ldots, 0)^T \in \mathbb{R}^n$ with 1 in the $i$th place and 0
elsewhere is a special case of $\mathbf{p}$ that assigns mass 1 to
the $i$-th observation. Namely, the influence function $\dot{S}_i$
is the directional derivative of $S(\mathbf{p})$ at $\mathbf{p}_0$
in the direction of $\mathbf{e}_i - \mathbf{p}_0$, which
essentially inspects the effect of an infinitesimal contamination
at the $i$th observation on the estimator.

\iffalse With the similar finite difference method, define the
second-order influence value
\begin{equation}
\ddot{S}_i  = \lim_{\epsilon \rightarrow 0^+}
\frac{\{S(\mathbf{p}_0 + \epsilon (\mathbf{e}_i - \mathbf{p}_0)) -
S(\mathbf{p}_0) \} \, - \, \{S(\mathbf{p}_0) -  S(\mathbf{p}_0 -
\epsilon (\mathbf{e}_i - \mathbf{p}_0))\} }{\epsilon^2}.
\label{S-ddot-i}
\end{equation}
\fi
The infinitesimal jackknife method \citep{Efron:1982} provides
an estimate of variance for $\hat{\delta}(\mathbf{x})$ given by
\begin{equation}
\hat{\mbox{var}}(\hat{\delta}(\mathbf{x}))  = \sum_{i=1}^n
\dot{S}_i^2 /n^2. \label{V-delta}
\end{equation}
Thus it remains to compute $\dot{S}_i.$

For a multinomial probability $\mathbf{p} = \left( p_i
\right)_{i=1}^n$, consider
\begin{equation}
S(\mathbf{p}) = \sum_{b=1}^B \left\{ \hat\delta_b (\mathbf{x})
\cdot \prod_{i=1}^n p_i^{N_{bi}} \right\}  = \sum_{b=1}^B
\frac{w_b(\mathbf{p}) \hat\delta_b (\mathbf{x})}{B}, \label{Sp}
\end{equation}
where $w_b(\mathbf{p}) = \prod_{i=1}^n (n p_i)^{N_{bi}}$ is the
ratio of the product of probabilities in $\mathbf{p}$ to that in
$\mathbf{p}_0.$ Note that the multinomial coefficient ${n \choose
N_{b1}, N_{b2}, \ldots, N_{bn}}$ does not show up in the summand
of (\ref{Sp}) since all $B = n^n$ choices of bootstrap samples are
listed in the sum, in which case some of the bootstrap samples can
be identical when ignoring the order of observations.

When
$$\mathbf{p} = \mathbf{p}_0 + \epsilon (\mathbf{e}_i -
\mathbf{p}_0) = \left[ \frac{1-\epsilon}{n}, \ldots,
\frac{1-\epsilon}{n}, \frac{1+ \epsilon (n-1)}{n},
\frac{1-\epsilon}{n}, \ldots, \frac{1-\epsilon}{n}\right]^T$$ with
$(1+ \epsilon (n-1))/n$ in the $i$-th position and
$(1-\epsilon)/n$ elsewhere, we have, letting $\epsilon \rightarrow
0^+$,
\begin{eqnarray*}
 w_b (\mathbf{p}) &= & (1-\epsilon)^{\sum_{i' \neq i} N_{bi'}}
\{1+ (n-1) \epsilon \}^{N_{bi}} \\
& \approx & ( 1 - \epsilon \sum_{i' \neq i} N_{bi'} ) \cdot
\left\{ 1 + (n-1) \epsilon N_{bi} \right\}; \\
&=& 1 - \epsilon \sum_{i' \neq i} N_{bi'}  + (n-1) \epsilon N_{bi}
+ O_p (\epsilon^2) \\
& \approx & 1 + n \epsilon (N_{bi}-1)
\end{eqnarray*}
where the second step uses the fact $(1 + x)^a \approx 1 + ax$ for
$|x| <1$ and any constant $a$ obtained from the bionomial series
and the fourth step ignores a second-order term of $\epsilon.$
Bringing $w_b (\mathbf{p})$ into (\ref{Sp}), it follows that
$$ S(\mathbf{p}_0 + \epsilon (\mathbf{e}_i -
\mathbf{p}_0)) \approx S(\mathbf{p}_0) + n \epsilon \sum_{b=1}^B
(N_{bi} -1) \hat{\delta} /B  = S(\mathbf{p}_0) + n \epsilon
\bar{Z}_i, $$ using the fact that $\sum_{i=1}^m (x_i -
\bar{x})(y_i - \bar{y}) = \sum_i (x_i - \bar{x})y_i$ for any $m$
real-valued pairs $(x_i, y_i).$  According to its definition in
(\ref{S-dot-i}), it is clear that $\dot{S}_i = n \bar{Z}_i.$
Bringing it back into (\ref{V-delta}) yields the needed result for
$\hat{V}$ in (\ref{SE-IJ}).  An alternative way of deriving
$\hat{V}_c$ is via H\'{a}jek projection as in \citet{Wager:2014}.
\hfill $\square$

In practice, a total of finite $B$ bootstrap sample is taken
instead, which makes $\hat{V}$ subject to additional Monte Carlo
noise. Following the bias correction step suggested by
\citet{Efron:2014}, let $\tilde{Z}_i = \sum_{b=1}^B Z_{bi}/B$
denote the $\bar{Z}_i$ value obtained from $B$ bootstrap samples,
as opposed to $\bar{Z}$ obtained from the ideal bootstrap. Thus
$\tilde{Z}_i$ has bootstrap mean $\bar{Z}_i$ and bootstrap
variance $ \nu_i^2 /B,$ where $\nu_i^2$ denote the bootstrap
variance of $Z_{bi}.$ It follows that
$$ E_{\mathcal{B}} \tilde{V} =  E_{\mathcal{B}} (\sum_{i=1}^n \tilde{Z}_i^2)
= \sum_{i=1}^n \bar{Z}_i^2 + \sum_{i=1}^n \nu_i^2 /B,$$ where
$E_{\mathcal{B}}(\cdot)$ denotes bootstrap expectation. Therefore,
a bias-corrected version for $\hat{V}$ is given by
\begin{equation}
\hat{V}_c = \hat{V} - \frac{1}{B} \sum_{i=1}^n \hat{\nu}_i^2,
\label{Vc-0}
\end{equation}
where $v_i^2$ is replaced with its estimate $\hat{\nu}_i^2 =
\sum_{b=1}^B (Z_{bi} - \tilde{Z}_i)^2/B.$ This gives the
expression in (\ref{SE-corrected-0}).

\citet{Wager:2014} further assumed that $N_{bi}$ and
$\hat{\delta}(\mathbf{x})$ in $Z_{bi}$ are approximately
independent. Under this assumption, we have
\begin{eqnarray*}
\nu_i^2 = \mbox{var}_{\mathcal{B}}(Z_{bi}) &=&
\mbox{var}_{\mathcal{B}}\left[ (N_{bi}-1)
\{\hat{\delta}_b(\mathbf{x}) - \hat{\delta}_b(\mathbf{x})\} \right] \\
&=& \mbox{var}_{\mathcal{B}}(N_{bi}) \cdot
\mbox{var}_{\mathcal{B}} (\hat{\delta}_b(\mathbf{x})) \\
&=& \frac{n-1}{n} \cdot \mbox{var}_{\mathcal{B}}
(\hat{\delta}_b(\mathbf{x})),
\end{eqnarray*}
where $\mbox{var}_{\mathcal{B}}(N_{bi}) = n (1/n) (1-1/n) =
(n-1)/n$. Thus $Z_{bi}$ becomes homoscedastic for different $i$
(i.e., with equal bootstrap variance). A natural estimate of
$v_i^2$ is
$$ \hat{v}_i^2 = \frac{n-1}{n} \cdot \frac{1}{B} \sum_{b=1}^B
\{\hat{\delta}_b (\mathbf{x}) - \hat{\delta}(\mathbf{x}) \}^2.$$
Bringing $\hat{v}_i^2$ into (\ref{Vc-0}) yields the bias-corrected
version in (\ref{SE-corrected}). This completes the proof. \hfill
$\blacksquare$

\subsection{Proof of Proposition \ref{prop-AMSE}}
\label{sec-proof-prop-AMSE} For the RFIT estimate
$\hat{\delta}(\mathbf{X}),$ rewrite $ E_{\mathbf{X}, \mathcal{D},
\mathcal{B}} \{\hat{\delta}(\mathbf{X}) - \delta(\mathbf{X}) \}^2$
as
\begin{equation}
 E_{\mathbf{X}, \mathcal{D}, \mathcal{B}} \left[ \left\{
\hat{\delta}(\mathbf{X}) - \bar{\delta}(\mathbf{X}; \mathcal{D})
\right\}
 + \left\{\bar{\delta}(\mathbf{X}; \mathcal{D}) -
\bar{\delta}(\mathbf{X}) \right\} + \left\{
\bar{\delta}(\mathbf{X}) - \delta (\mathbf{X}) \right\} \right]^2.
\label{supp-rewrite}
\end{equation}
By the definitions of $\{\bar{\delta}(\mathbf{X}; \mathcal{D}),
\bar{\delta}(\mathbf{X})\}$, each of the cross-product terms
amounts to 0, which leads immediately to (\ref{AMSE-RFIT}).

For the SR estimate $\hat{\delta}(\mathbf{X})$ in (\ref{ITE-SR}),
\begin{eqnarray*}
\mbox{AMSE} &=& E_{\mathbf{X}, \mathcal{D}, \mathcal{B}} \left[
\left\{ \hat{\mu}_1 (\mathbf{X}) - \mu_1(\mathbf{X}) \right\}
 - \left\{ \hat{\mu}_0 (\mathbf{X}) - \mu_0(\mathbf{X}) \right\}
\right]^2 \\
&=& E_{\mathbf{X}, \mathcal{D}, \mathcal{B}} \left\{ \hat{\mu}_1
(\mathbf{X}) - \mu_1(\mathbf{X}) \right\}^2 + E_{\mathbf{X},
\mathcal{D}, \mathcal{B}} \left\{ \hat{\mu}_0 (\mathbf{X}) -
\mu_0(\mathbf{X}) \right\}^2 \\
&& - 2 E_{\mathbf{X}, \mathcal{D}, \mathcal{B}} \left\{
\hat{\mu}_1 (\mathbf{X}) - \mu_1(\mathbf{X}) \right\} \left\{
\hat{\mu}_0 (\mathbf{X}) - \mu_0(\mathbf{X}) \right\}
\end{eqnarray*}
The first two terms can be rewritten in a similar manner to
(\ref{supp-rewrite}), with all cross-product terms being 0. For
the last term, note that $\hat{\mu}_1 (\mathbf{X})$ and
$\hat{\mu}_0 (\mathbf{X})$ are based on the separate parts of data
$\mathcal{D}$ and hence are independent given $\mathcal{D}.$ It
follows that
\begin{eqnarray*}
&& E_{\mathbf{X}, \mathcal{D}, \mathcal{B}} \left\{ \hat{\mu}_1
(\mathbf{X}) - \mu_1(\mathbf{X}) \right\} \left\{ \hat{\mu}_0
(\mathbf{X}) - \mu_0(\mathbf{X}) \right\} \\
&=& E_{\mathbf{X}}
\left[ E_{\mathcal{D}, \mathcal{B}} \left\{ \hat{\mu}_1
(\mathbf{X}) - \mu_1(\mathbf{X}) \right\} \cdot  E_{\mathcal{D},
\mathcal{B}} \left\{ \hat{\mu}_0 (\mathbf{X}) - \mu_0(\mathbf{X})
\right\}  \, | \, \mathbf{X} \right] \\
&=& E_{\mathbf{X}} \left[\{\bar{\mu}_1(\mathbf{X}) -
\mu_1(\mathbf{X})\} \{\bar{\mu}_0(\mathbf{X}) -
\mu_0(\mathbf{X})\} \right].
\end{eqnarray*}
The proof is completed.  \hfill $\blacksquare$

\section{Additional Numerical Results}
\label{sec-SM-numerical}

We have implemented RFIT with R \citep{R:2017} and an R package \textbf{RFIT} is underway. This section contains additional numerical results that we have omitted from the manuscript due to page limitation and presents a variable description for the headache data set used in Section \ref{sec-example}.

\subsection{Bias in Section \ref{sec-simulation-RFIT-SR}}
\label{sec-bias}

As we have discussed in Section \ref{sec-MSE}, SR may suffer more
from the bias problem. This perspective can be further explored by
plotting the predicted ITE $\hat{\delta}$ versus the true ITE
$\delta$.

\begin{figure}[h]
\centering
  \includegraphics[scale=0.55, angle=0]{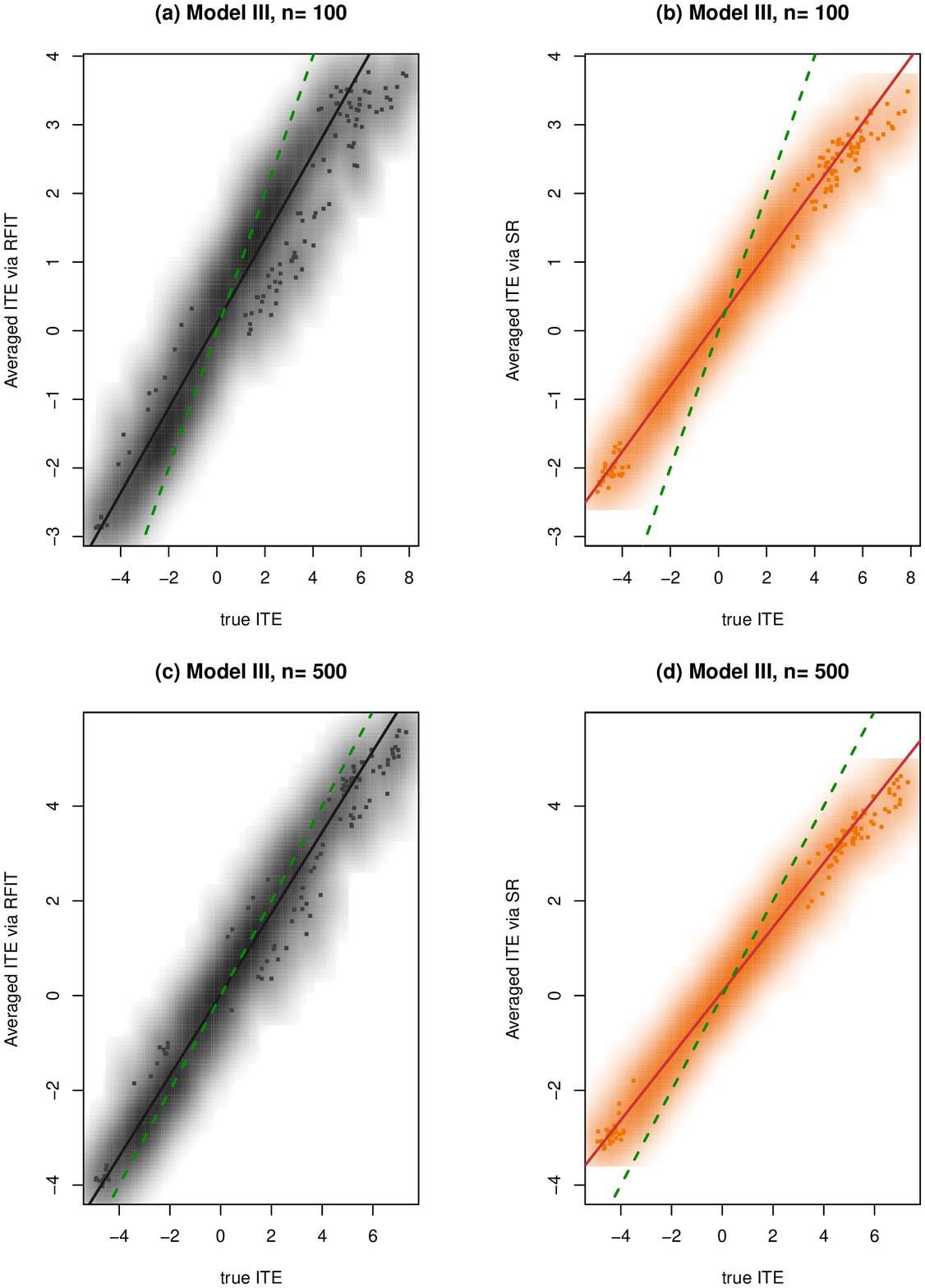}
  \caption{Smoothed scatterplot of the predicted ITE $\hat{\delta}(\mathbf{x})$
averaged over 200 runs versus the true ITE $\delta(\mathbf{x})$:
RFIT (in {\color{gray}{gray}} and SR (in
{\color{orange}{orange}}). Predictions are based on a test sample
of $n'=2000$ generated from Model III in (\ref{Model-III}). Two
sample sizes $n=100$ and $n=500$ are considered for the training
sample. The {\color{green}{green}} dashed line highlights the
reference line $y=x$ with perfect prediction. \label{fig-I}}
\end{figure}

With the same settings as in Section \ref{sec-simulation-RFIT-SR},
Figure \ref{fig-I} provides such a plot for Model III in
(\ref{Model-III}) as an example. Here the predicted
$\hat{\delta}(\mathbf{x})$ for each of $n'=2000$ observations in a
test sample are averaged over 200 simulation runs, yielding an
estimate of $E_{\mathcal{D}, \mathbf{B}}
(\hat{\delta}(\mathbf{x}))$ or $\bar{\delta}(\mathbf{x})$ in
Proposition \ref{prop-AMSE}, and then plotted versus the true ITE
$\delta(\mathbf{x}).$ The smoothed scatterplot is used because of
the large sample size and hence many overlapping dots. The left
panels (a) \& (c) have RFIT predictions on the vertical axis while
the right panels (b) \& (d) have SR predictions on the vertical
axis. Each panel is superimposed with the LS fitted line, as well
as the reference line $y=x$ with perfect prediction. Panels (a) \&
(b) are based on same training data sets of size $n=100$; for
comparison purposes, we have made the ranges for the vertical axis
the same; similarly for Panels (c) \& (d).

Referring to the line $y=x$, both RFIT and SR are subject to bias in predicting high and low values of ITE. They tend to cut the hill (predicting high ITE values lower) and fill the valley (predicting low ITE values high). However, this bias problem is more prominent for SR than for RFIT. With increased sample sizes ($n=500$), the bias is diminishing for RFIT but remains substantial for SR. For the same reason, the range of SR estimates is narrower than that of RFIT, resulting in smaller variances though. Similar patterns can be observed with other models and configurations, for which the plots are omitted here. Another potential source of bias for RFIT stems from unbalanced covariates within terminal nodes. Although the treatment assignment is randomized for the entire study, the balance among covariate distributions might be lost as the tree grows. Thus, additional adjustment for confounders within terminal nodes might further improve RFIT.

\subsection{Variable Description for the Headache Data}
\label{sec-variable-description}

The headache data set that was used for illustration of RFIT in
Section \ref{sec-example} contains 21 variables on 301 subjects
who completed the trial. There are three subjects with some
missing data, which are imputed with random forests (see R pacakge
\textbf{missForest}; Stekhoven and Buehlmann, 2012). A brief
variable description for the final compiled data is provided in
Table \ref{tblS1}, where the variable names are consistent with
those in the original data file from \citep{Vickers:2004} and
Vickers (2006).
% The final data set, named \texttt{headache}, is included in our R Package \textbf{RFIT}.

\renewcommand{\tabcolsep}{6.pt}
\renewcommand{\arraystretch}{1.1}
\renewcommand{\baselinestretch}{1.0}
\begin{table}[h]
\caption{Variable Description for the Headache Data.}
\vspace{.4in}
% \rule{5.5in}{.03cm}
\centering
\begin{tabular}{ll} \hline \hline
Name & Description \\
\hline \texttt{id} &  Patient ID code \\
\texttt{diff} & Difference in headache severity score between one year follow-up \\
& and baseline, i.e., (\texttt{pk5} - \texttt{pk1}) \\
\texttt{group} &   0 is control; 1 is acupuncture \\ \hline
\texttt{age}  & Age \\
\texttt{sex}  & sex: 0 male; 1 female \\
\texttt{migraine} & Migraine: 0 No and 1 Yes \\
\texttt{chronicity} &  Chronicity \\
\texttt{pk1} &  Severity score at baseline \\
\texttt{f1} &  Headache frequency at baseline \\
\texttt{pf1} & Baseline SF36 physical functioning \\
\texttt{rlp1} &    Baseline SF36 role limitation physical \\
\texttt{rle1} &    Baseline SF36 role limitation emotional \\
\texttt{ef1} & Baseline SF36 energy fatigue \\
\texttt{ewb1} &    Baseline SF36 emotional well being \\
\texttt{sf1} & Baseline SF36 social functioning \\
\texttt{p1} &  Baseline SF36 pain \\
\texttt{gen1} &    Baseline SF36 general health \\
\texttt{hc1} & Baseline SF36 health change \\
\texttt{painmedspk1} & MQS at baseline \\
% \texttt{prophylacticd1} & dosage weight of prophylactic medication at baseline \\
\texttt{prophmqs1}  & MQS of prophylactic medication at baseline \\
\texttt{allmedsbaseline} & Total MQS at baseline \\ \hline
\end{tabular}
% \rule{8.2in}{.03cm}
\label{tblS1}
\end{table}

\iffalse
\section{R Package \textbf{RFIT}}
\label{sec-R-package-RFIT}
We put together an R package, named \textbf{RFIT}, for the proposed method.
\fi

\vspace{.3in}
\noindent
\textbf{Additional References}
\begin{description}
\item R Core Team (2017). \textit{R: A language and environment for
statistical computing}. R Foundation for Statistical Computing,
Vienna, Austria. URL~\url{https://www.R-project.org/}.

\item Stekhoven, D.~J.~and Buehlmann, P.~(2012).
\textbf{missForest} -- nonparametric missing value imputation for
mixed-type data. \textit{Bioinformatics}, \textbf{28}: 112--118.

\item Vickers, A.~J.~(2006). Whose data set is it anyway? Sharing
raw data from randomized trials. \textit{Trials}, \textbf{7}: 15.
doi: 10.1186/1745-6215-7-15.
\end{description}


\begin{thebibliography}{99}
% \iffalse
\expandafter\ifx\csname
natexlab\endcsname\relax\def\natexlab#1{#1}\fi
\expandafter\ifx\csname url\endcsname\relax
  \def\url#1{\texttt{#1}}\fi
\expandafter\ifx\csname
urlprefix\endcsname\relax\def\urlprefix{URL }\fi
% \fi

\bibitem[Ballman(2015)]{Ballman:2015}
Ballman, K.~V.~(2015). Biomarker: Predictive or Prognostic?
\textit{Journal of Clinical Oncology}, \textbf{33}: 3968--3971.


\iffalse
\bibitem[Berger, Wang, and Shen(2014)]{Berger:2014}
Berger, J., Wang, X., and Shen, L.~(2014). A Bayesian approach to
subgroup identification. \textit{Journal of Biopharmaceutical
Statistics}, \textbf{24}: 110--129.
\fi

\iffalse
\bibitem[Bien, Taylor, and Tibshirani(2013)]{Bien:2013}
Bien, J., Taylor, J., and Tibshirani, R.~(2013). A lasso for
hierarchical interactions. \textit{The Annals of Statistics},
\textbf{41}: 1111--1141. \fi


\bibitem[Breiman(1999)]{Breiman:1999}
Breiman, L.~(1999). Using adaptive bagging to debias regressions.
Technical Report \# 547, Department of Statistics, University of
California at Berkely.

\bibitem[Breiman(2001)]{Breiman:2001}
Breiman, L.~(2001). Random Forests. \textit{Machine Learning},
\textbf{45}: 5--32.


\bibitem[Breiman et al.(1984)]{Breiman:1984}
Breiman, L., Friedman, J., Olshen, R., and Stone, C.~(1984).
\textit{Classification and Regression Trees}. Belmont, CA:
Wadsworth International Group.

\bibitem[Brent(1973)]{Brent:1973}
Brent, R.~(1973). \textit{Algorithms for Minimization without
Derivatives}. Englewood Cliffs, NJ: Prentice-Hall.


\iffalse
\bibitem[Cai et al.(2011)]{Cai:2011}
Cai, T., Tian, L., Wong, P., and Wei, L.~J.~(2011). Analysis of
randomized comparative clinical trial data for personalized
treatment selections. \textit{Biostatistics}, \textbf{12}:
270--282.
\fi

\bibitem[Dusseldorp and van Mechelen(2014)]{Dusseldorp:2014}
Dusseldorp, E.~and van Mechelen, I.~(2014). Qualitative interaction trees: a tool to identify qualitative treatment-subgroup interactions.
\textit{Statistics in Medicine}, \textbf{33}: 219--237.

\bibitem[Efron(1982)]{Efron:1982}
Efron, B.~(1982). The jackknife, the bootstrap and ohter
resampling plans. CBMS-NSF Regional COnference Series in Applied
Mathematics 38. Philadelphia, PA: Society for Industrial and
Applied Mathematics (SIAM).

\bibitem[Efron(2014)]{Efron:2014}
Efron, B.~(2014). Estimation and accuracy after model selection
(with discussion). \textit{Journal of the American Statistical
Association}, \textbf{109}: 991--1007.

\iffalse
\bibitem[Efron and Stein(1981)]{Efron:1981}
Efron, B.~and Stein, C.~(1981). The jackknife estimate of
variance. \textit{The Annals of Statistics},
\textbf{9}(3):586--596.
\fi

\bibitem[Foster, Taylor, Ruberg(2011)]{Foster:2011}
Foster, J.~C., Taylor, J.~M.~C., and Ruberg, S.~J.~(2011).
Subgroup identification from randomized clinical trial data.
\textit{Statistics in Medicine}, \textbf{30}: 2867--2880.

\bibitem[Friedman(1991)]{Friedman.1991}
Friedman, J.~H.~(1991). Multivariate adaptive regression splines.
\textit{Annals of Statistics}, \textbf{19}: 1--67.

\iffalse
\bibitem[Heckman et al.(1998)]{Heckman:1998}
Heckman, J.~J., Hidehiko I., Smith, J., and Petra Todd, P.~(1998).
Characterizing selection bias using experimental data.
\textit{Econometrica}, \textbf{66}: 1017--1098.
\fi

\bibitem[Imai and Ratkovic(2013)]{Imai:2013}
Imai, K.~and Ratkovic, M.~(2013). Estimating treatment effect
heterogeneity in randomized program evaluation. \textit{The Annals
of Applied Statistics}, \textbf{7}: 443--470.

\bibitem[Laber and Zhao(2015)]{Laber:2015}
Laber, E.~B.~and Zhao, Y.~Q.~(2015). Tree-based methods for
individualized treatment regimes. \textit{Biometrika},
\textbf{102}: 501--514.

\bibitem[LeBlanc and Crowley(1993)]{LeBlanc.1993}
LeBlanc, M.~and Crowley, J.~(1993). Survival trees by goodness of
split. \textit{Journal of the American Statistical Association},
\textbf{88}: 457--467.

\bibitem[Liaw and Wiener(2002)]{Liaw:2002}
Liaw, A.~and Wiener, M.~(2002). Classification and Regression by randomForest. \textit{R News}, \textbf{2}/3: 18--22.

\bibitem[Lipkovich, Dmitrienko, and D'Agostino(2016)]{Lipkovich:2017}
Lipkovich, I., Dmitrienko, A., D'Agostino, R.~B.~(2017). Tutorial
in biostatistics: data-driven subgroup identification and analysis
in clinical trials. \textit{Statistics in Medicine}, \textbf{36}:
136--196.

\bibitem[Lipkovich et al.(2011)]{Lipkovich:2011}
Lipkovich, I., Dmitrienko, A., Denne, J., and Enas, G.~(2011).
Subgroup identification based on differential effect search
(SIDES): a recursive partitioning method for establishing response
to treatment in patient subpopulations. \textit{Statistics in
Medicine}, \textbf{30}: 2601--2621.

\bibitem[Loh, He, and Man(2015)]{Loh:2015}
Loh, W.-Y., He, X., and Man, M.~(2015). A regression tree approach
to identifying subgroups with differential treatment effects.
\textit{Statistics in Medicine}, \textbf{34}: 1818--1833.

\bibitem[Murphy(2003)]{Murphy:2003}
Murphy, S.~A.~(2003). Optimal dynamic treatment regimes (with
discussion). \textit{Journal of the Royal Statistical Society,
Series B}, \textbf{65}: 331--366.

\bibitem[Neyman(1923)]{Neyman:1923}
Neyman, J.~(1923). On the application of probability theory to
agricultural experiments. \textit{Essay on Principles}, Section 9.
\textit{Statistical Science}, \textbf{5}: 465--472, 1990.
Translated by Dorota M.~Dabrowska and Terence P.~Speed.


\bibitem[R Core Team(2017)]{R:2017}
R Core Team (2017). \textit{R: A language and environment for
statistical computing}. R Foundation for Statistical Computing,
Vienna, Austria. URL~\url{https://www.R-project.org/}.


\bibitem[Rosenbaum and Rubin(1983)]{Rosenbaum:1983}
Rosenbaum, P.~R.~and Rubin, D.~B.~(1983). The central role of the
propensity score in observational studies for causal effects.
\textit{Biometrika}, \textbf{70}: 41--55.


\bibitem[Rubin(1974)]{Rubin:1974}
Rubin, D.~B.~(1974). Estimating Causal Effects of Treatments in
Randomized and Nonrandomized Studies. \textit{Journal of
Educational Psychology}, \textbf{66}: 688--701.


\bibitem[Rubin(2005)]{Rubin:2005}
Rubin, D.~B.~(2005). Causal inference using potential outcomes:
Design, modeling, decisions. \textit{Journal of the American
Statistical Association}, \textbf{100}: 322--331.

\bibitem[Shen and He(2015)]{Shen:2015}
Shen, J.~and He, X.~(2015). Inference for Subgroup Analysis with a
Structured Logistic-Normal Mixture Model. \textit{Journal of the
American Statistical Association}, \textbf{110}: 303--312.

\bibitem[Su et al.(2012)]{Su:2012}
Su, X., Kang, J., Fan, J., Levine, R., and Yan, X.~(2012).
Facilitating score and causal inference trees for large
observational data. \textit{Journal of Machine Learning Research},
\textbf{13}: 2955--2994.

\bibitem[Su et al.(2009)]{Su:2009}
Su, X., Tsai, C.-L., Wang, H., Nickerson, D. M., and Li,
B.~(2009). Subgroup analysis via recursive partitioning.
\textit{Journal of Machine Learning Research}, \textbf{10}:
141--158.

\bibitem[Utgoff, Berkman, and Clouse(1997)]{Utgoff:1997}
Utgoff, P.~E.,  Berkman, N.~C., and Clouse, J.~A.~(1997). Decision
tree induction based on efficient tree restructuring.
\textit{Machine Learning}, \textbf{29}: 5--44.

\bibitem[Wager, Hastie, and Efron(2014)]{Wager:2014}
Wager, S., Hastie, T., and Efron, B.~(2014). Confidence intervals
for random forests: The jackknife and the infinitesimal jackknife.
\textit{Journal of the Machine Learning Research}, \textbf{15}:
1625--1651.

\bibitem[Wager and Athey(2016)]{Wager:2016}
Wager, S.~and Athey, S.~(2016). Estimation and inference of
heterogeneous treatment effects using random forests.
\textit{arXiv preprint}, arXiv:1510.04342v3.


\bibitem[van der Laan, Polley, Hubbard(2006)]{van der Laan:2006}
van der Laan, M., Polley, E.,  and Hubbard, A.~(2007). Super
Learner. \textit{Statistical Applications in Genetics and
Molecular Biology},  \textbf{6}(1).


\iffalse
\bibitem[Vickers(2006)]{Vickers:2006}
Vickers, A.~J.~(2006). Whose data set is it anyway? Sharing raw
data from randomized trials. \textit{Trials}, \textbf{7}: 15. doi:
10.1186/1745-6215-7-15.
\fi

\bibitem[Vickers et al.(2004)]{Vickers:2004}
Vickers, A.~J., Rees, R.~W., Zollman, C.~E., McCarney, R., Smith,
C., Ellis, N., Fisher, P., and Van Haselen, R.~(2004). Acupuncture
for chronic headache in primary care: large, pragmatic, randomised
trial. \textit{British Medical Journal, Primary Care},
\textbf{328}: 744--749. % doi:10.1136/bmj.38029.421863.EB.

\bibitem[Zhang et al.(2012)]{Zhang:2012}
Zhang, B., Tsiatis, A.~A., Davidian, M., Zhang, M., and Laber, E.~(2012).
Estimating optimal treatment regimes from a classification perspective.
\textit{STAT}, \textbf{1}: 103--114.

\end{thebibliography}
\end{document}